\documentclass[letterpaper]{article} 
\usepackage{aaai25}
\usepackage{times}  
\usepackage{helvet}  
\usepackage{courier}  
\usepackage[hyphens]{url}  
\usepackage[colorlinks=false, breaklinks=true]{hyperref}
\usepackage{graphicx} 
\usepackage{natbib}  
\usepackage{caption} 
\usepackage{algorithm}
\usepackage{algorithmic}

\usepackage{newfloat}
\usepackage{listings}
\DeclareCaptionStyle{ruled}{labelfont=normalfont,labelsep=colon,strut=off} 
\floatstyle{ruled}
\newfloat{listing}{tb}{lst}{}
\floatname{listing}{Listing}
\usepackage{graphicx}
\usepackage{booktabs}
\usepackage{multirow}
\usepackage{array}
\usepackage{arydshln} 
\usepackage{xcolor}
\usepackage{multirow} 
\definecolor{DeepBlue}{rgb}{0.0, 0.0, 0.55}
\usepackage{longtable}
\usepackage{tcolorbox}
\usepackage{bm}
\usepackage{float}
\usepackage{booktabs} 
\usepackage{makecell}
\usepackage{amsmath, amsfonts, amsthm, amssymb}
\usepackage{soul}

\newcommand*\samethanks[1][\value{footnote}]{\footnotemark[#1]}

\title{InstructionBench: An Instructional Video Understanding Benchmark}

\author {
    Haiwan Wei\textsuperscript{1}\thanks{Equal contribution},
    Yitian Yuan\textsuperscript{2}\samethanks,
    Xiaohan Lan\textsuperscript{2},
    Wei Ke\textsuperscript{1},
    Lin Ma\textsuperscript{2}\thanks{Project leader}
}

\affiliations{
    \textsuperscript{1}School of Software Engineering, Xi'an Jiaotong University \\
    \textsuperscript{2}Meituan Inc. \\
}

\begin{document}

\maketitle

\begin{abstract}

Despite progress in video large language models (Video-LLMs), research on instructional video understanding, crucial for enhancing access to instructional content, remains insufficient. To address this, we introduce InstructionBench, an Instructional video understanding Benchmark, which challenges models' advanced temporal reasoning within instructional videos characterized by their strict step-by-step flow. Employing GPT-4, we formulate Q\&A pairs in open-ended and multiple-choice formats to assess both Coarse-Grained event-level and Fine-Grained object-level reasoning. Our filtering strategies exclude questions answerable purely by common-sense knowledge, focusing on visual perception and analysis when evaluating Video-LLM models. The benchmark finally contains 5k questions across over 700 videos. We evaluate the latest Video-LLMs on our InstructionBench, finding that closed-source models outperform open-source ones. However, even the best model, GPT-4o, achieves only 53.42\% accuracy, indicating significant gaps in temporal reasoning. To advance the field, we also develop a comprehensive instructional video dataset with over 19k Q\&A pairs from nearly 2.5k videos, using an automated data generation framework, thereby enriching the community's research resources.
All data are available at \url{https://huggingface.co/datasets/sunwhw/InstructionBench}.

\end{abstract}

\section{Introduction}

In recent years, significant advances have been made in the realm of Video Large Language Models (Video-LLMs). Notable models~\cite{li2024mvbench,zhang2024llavanextvideo,cheng2024videollama} have pushed the boundaries of video understanding capabilities in multiple dimensions. Concurrently, specialized datasets and benchmarks have emerged to support the development of Video-LLMs. Training datasets for Video-LLMs generally fall into two categories: pre-training and instruction tuning. Pre-training~\cite{bain2021frozen,chen2024panda} datasets offer diverse video content to establish core visual-language alignments, while instruction tuning datasets~\cite{maaz2024videochatgptdetailedvideounderstanding,li2024mvbench} enhance model interaction through detailed inquiries and instructions. Additionally, several comprehensive benchmarks~\cite{ning2023video,mangalam2023egoschemadiagnosticbenchmarklongform,fu2024video} have been developed to provide robust evaluation methods for Video-LLMs across various tasks.

However, specialized research in instructional video understanding remains limited, hindering efficient instructional content acquisition. This domain requires models to parse and reason about procedural knowledge, identifying steps and logical sequences. Existing datasets~\cite{zhou2018towards, zala2023hierarchicalvideomomentretrievalstepcaptioning} offer instructional scenarios but lack complex task designs for advanced temporal reasoning.~\cite{ren2024timechat} enhances instructional video datasets with fill-in-the-blank tasks but remains limited in instruction format diversity and temporal reasoning capabilities. Moreover, current datasets have not adequately addressed the impact of common-sense knowledge in instructional videos, potentially causing Video-LLMs to rely on textual hints rather than genuine visual analysis.

\begin{figure*}[t]
    \centering
    \includegraphics[width=0.9\textwidth]{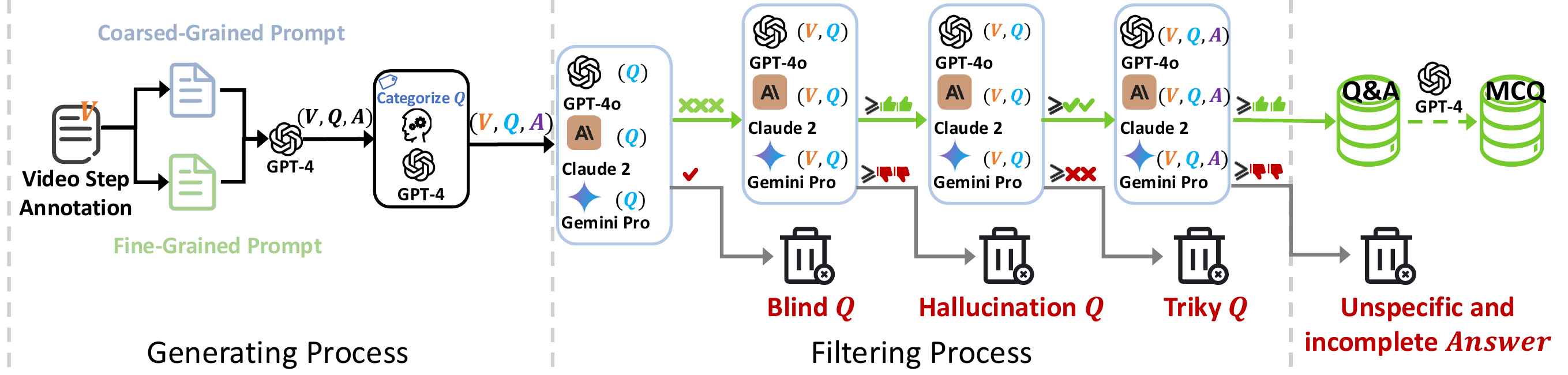}
    \caption{The overview of the construction process of InstructionBench. In the filtering phase, green markers indicate no issues, while red markers indicate errors, especially during the ``Blind Q'' filtering phase, where questions answered correctly without viewing the video should be discarded.}
    \label{fig:data_construction_process}
\end{figure*}

Based on these considerations, we introduce a benchmark named \textbf{InstructionBench}, focusing on instructional video understanding, with a particular emphasis on temporal reasoning. Moreover, we construct a instructional video training dataset. Both the benchmark and the training dataset utilize 4 source instructional video datasets with diverse instructional scenarios and high-quality manual annotations, through an automated Q\&A generation and filtering framework.

Given the step-by-step nature of instructional videos, we craft a question design framework focusing on the video's time sequence, dividing it into Coarse-Grained (event-level) and Fine-Grained (object-level). Coarse-Grained questions focus on action sequences and specific activities, requiring models to recognize and sequence key actions using temporal reasoning. Fine-Grained questions focus on objects but still relate to specific actions, requiring models to identify objects and their connection to the action timeline. Based on this framework, we design detailed prompts for both levels to engage GPT-4~\cite{achiam2023gpt} for generating Q\&A pairs.

The overview of the construction process of InstructionBench is shown in Figure~\ref{fig:data_construction_process}. Firstly, to generate open-ended Q\&A pairs for our InstructionBench, we organize the video's step annotations in temporal order, which include specific descriptions and timestamps of actions/events. Subsequently, these annotations are paired with prompts of different granularity, enabling GPT-4 to produce distinct sets of Q\&A pairs at both the Coarse-Grained and Fine-Grained levels. Meanwhile, by employing GPT-4 along with some human assistance, we conduct a more detailed categorization of the generated questions. Examples of different question types are shown in Table~\ref{tab:new_task_dimension}.

\begin{figure*}[t]
    \centering
    \includegraphics[width=0.8\textwidth]{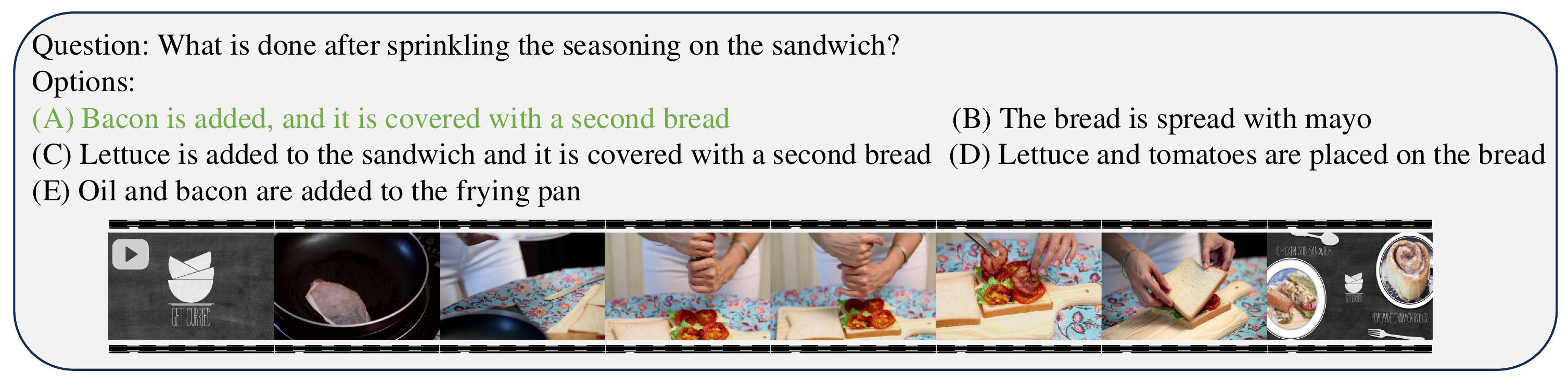}
    \caption{An example of Future Step Prediction task from the Coarsed-Grained~(event-level) category in our proposed InstructionBench, where the green-highlighted option is the correct one. This video is from the YouCook2~\cite{zhou2018towards}, and the link is \url{https://www.youtube.com/watch?v=4eWzsx1vAi8}, with the clue related to this question appearing around 03:05 in the video.}
    \label{fig:example_case1}
\end{figure*}

To ensure quality of Q\&A pairs in our InstructionBench, we also implement meticulous filtering strategies. For instance, to filter out questions answerable solely with common-sense knowledge, we provide questions without step annotations to multiple LLM assistants (GPT-4o~\cite{openai2024gpt4o}, Gemini Pro~\cite{team2023gemini}, and Claude 2~\cite{claude}) and remove questions any assistant answers correctly. This strategy emphasizes visual reasoning over common-sense understanding. We also filter out questions unrelated to video annotations and remove tricky or obscure questions and unclear answers. Ultimately, InstructionBench includes 5k Q\&A pairs covering over 700 videos. In order to facilitate further evaluation, we also utilize GPT-4 to convert the open-ended Q\&A pairs to Multiple-Choice (MC) format. One example of the MC-formatted question answering is shown in Figure~\ref{fig:example_case1}. A comprehensive breakdown of data statistics and question type distributions is provided in Figure~\ref{fig:data_statistics_and_task_type_distribution}.

By referring to the aforementioned automated data generation and filtering process, we also generate a instructional video training dataset, with over 19k Q\&A pairs spanning nearly 2.5k videos, thus enhancing the training resources available in the instructional video understanding domain.


Based on InstructionBench, we evaluate various open-source and closed-source models. Results indicate closed-source models, particularly GPT-4o~\cite{openai2024gpt4o} (48.66\% accuracy), Gemini Pro Vision~\cite{reid2024gemini} (42.28\%) and GPT-4V~\cite{gpt4v}(41.84\%) outperform open-source ones by 16-22\% using 8 frames. Increasing to 16 frames significantly improves closed-source model performance, with GPT-4o reaching 53.42\% accuracy. However, current video large language models still struggle with temporal reasoning in instructional videos.

The main contributions of this paper are:
\begin{itemize}
    \item We construct a novel benchmark, InstructionBench, for evaluating Video-LLMs on temporal reasoning in instructional video scenarios.
    \item We develop an automated Q\&A generation framework that unifies the processes of generating and filtering Q\&A pairs, previously utilized in the construction of InstructionBench. Based on it, we create an instructional video training dataset of nearly 19k Q\&A pairs.
    \item We conduct comprehensive evaluations of a range of open-source and closed-source Video-LLMs in InstructionBench, demonstrating that current models notably underperform on temporal reasoning with instructional videos.
\end{itemize}

\begin{table*}[!t]
    \centering
    \scalebox{0.65}{
        \begin{tabular}{c c p{13cm}}
        \toprule
        \textbf{Task Type} & \textbf{Question Type} & \textbf{Question Sample} \\
        \midrule
        \multirow{5}{*}{\textbf{Coarse-Grained (event-level)}} 
        & Future Step Prediction & After mincing the dough in the divider, what was the next step taken in the bread preparation? \\
        \cdashline{2-3}\noalign{\vskip 0.5ex}
        & Past Step Recall & Prior to placing the dough in the dough mixer, what did I do? \\
        \cdashline{2-3}\noalign{\vskip 0.5ex}
        & Specific Step Recognition & What part in the procedure did shaping the dough into balls come? \\
        \cdashline{2-3}\noalign{\vskip 0.5ex}
        & Intermediate Steps Recognition & What was done between chopping the potatoes and sauteing them? \\
        \cdashline{2-3}\noalign{\vskip 0.5ex}
        & Step Sequencing & Can you tell me the sequence of events while I was prepping and washing the dishes? \\
        \midrule
        \multirow{3}{*}{\textbf{Fine-Grained (object-level)}} 
        & Object Existence in Steps & What steps involved the tomato after I got it? \\
        \cdashline{2-3}\noalign{\vskip 0.5ex}
        & Object Attribute Recognition & Where was the measuring spoon placed after use? \\
        \cdashline{2-3}\noalign{\vskip 0.5ex}
        & Object Interaction Recognition & On which object is the toast sesame oil placed after use? \\
        \bottomrule
        \end{tabular}
    }
    \caption{Question types and examples in our proposed InstructionBench.}
    \label{tab:new_task_dimension}
\end{table*}

\section{Related Work}
\subsubsection{Video Understanding with Large Language Models (Video-LLMs).} Video-LLMs mainly evolved from Image-LLMs, as seen in works like ~\cite{yang2022zero, chen2023videollm}. These models encoded multiple video frames as individual images processed through an LLM, relying solely on the LLM's temporal and contextual processing. Subsequently, models began incorporating multi-frame temporal modeling to capture sequential video information. Early models like ~\cite{luo2023valley, maaz2024videochatgptdetailedvideounderstanding} used simple pooling strategies. Later models introduced temporal encoding modules like ~\cite{li2023videochat}, while recent approaches ~\cite{lin2023video, li2024mvbench, cheng2024videollama} utilize video encoders capturing temporal dynamics more effectively. While earlier efforts primarily addressed short video understanding, there are also some methods like~\cite{li2023llama, zhang2024llavanextvideo} now focus on long video comprehension using token compression and linear scaling.


\subsubsection{Datasets for Video-LLMs.} Current methods for training Video-LLMs primarily use two types of datasets: pretraining and instruction tuning. Pretraining datasets, such as ~\cite{zellers2021merlot, xue2022hdvila}, aim for visual-language alignment. However, ~\cite{bain2021frozen} found that ASR-generated text annotations often fail to semantically align with videos, highlighting the importance of quality over quantity. Later datasets, such as~\cite{nagrani2022learning, wang2023internvid, chen2024panda}, focus more on this alignment. Instruction tuning datasets like ~\cite{li2023llama, song2024moviechat, maaz2024videochatgptdetailedvideounderstanding, li2024mvbench} come from sources like YouTube, ~\cite{caba2015activitynet}, movies or combined. However, there is a lack of training data specifically for instructional videos. Traditional instructional video datasets like ~\cite{zhou2018towards, zhukov2019cross, tang2019coin, zala2023hierarchical} do not meet the expressive and reasoning needs of modern Video-LLMs. Although~\cite{ren2024timechat} enhances existing instructional datasets with fill-in-the-blank tasks, it remains limited by constrained Q\&A formats and insufficient temporal reasoning.

\subsubsection{Benchmarks for Video-LLMs.}
Early Video-LLMs' evaluation were based on established basic Q\&A assessments, such as ~\cite{xu2017video, jang2017tgif, xu2017video,  lei2018tvqa, yu2019activitynet, li2020hero, xiao2021next, wu2024star}. However, these benchmarks are limited due to various constraints, such as the short average duration of videos, a narrow range of video domains covered, and question formats lacking variety, among others. Recent benchmarks like ~\cite{ning2023video, li2023seed, li2024mvbench, li2024videovista, fu2024video, wang2024lvbench, liu2024tempcompass} offer comprehensive evaluations but often lack a focus on instructional videos. Even though ~\cite{ning2023video} touches upon tasks like summarization within the instructional video dataset ~\cite{zhou2018towards}, there remains a lack of focus on heightened temporal reasoning abilities that are crucial in the context of instructional videos, which inherently possess a step-by-step nature and logical sequences. To address this gap, we introduce a specialized benchmark specifically designed to evaluate temporal reasoning in instructional video scenarios.

\begin{figure*}[!t]
    \centering
    \begin{minipage}[c]{0.35\textwidth}
        \centering
        \setlength{\tabcolsep}{0.9pt}
        \renewcommand{\arraystretch}{1.2}
        \small
        \begin{tabular}{lc}
            \toprule
            \textbf{Category} & \textbf{Size} \\
            \midrule
            Task Classes & 8 \\
            \hspace{3mm}- Coarsed-Grained Temporal Reasoning & 5 \\
            \hspace{3mm}- Fine-Grained Temporal Reasoning & 3 \\
            Video & - \\
            \hspace{3mm}- Source Datasets: (YouCook2, HiREST, \\ \hspace{6mm}Ego4D Goal-Step, Ego-Exo4D) & \multirow{2}{*}{\vspace{2mm}4} \\
            \hspace{3mm}- Videos & 713 \\
            \hspace{3mm}- Video Clips & 932 \\
            \hspace{3mm}- Average Duration & 282.95s \\
            Average Question Length & 11.66 \\
            Average Option Length & 10.92 \\
            Option Numbers & 5 \\
            Total Samples & 5,000 \\
            Total Questions & 5,000 \\
            \bottomrule
        \end{tabular}
    \end{minipage}%
    \hfill
    \begin{minipage}[c]{0.6\textwidth}
        \centering
        \includegraphics[width=\linewidth]{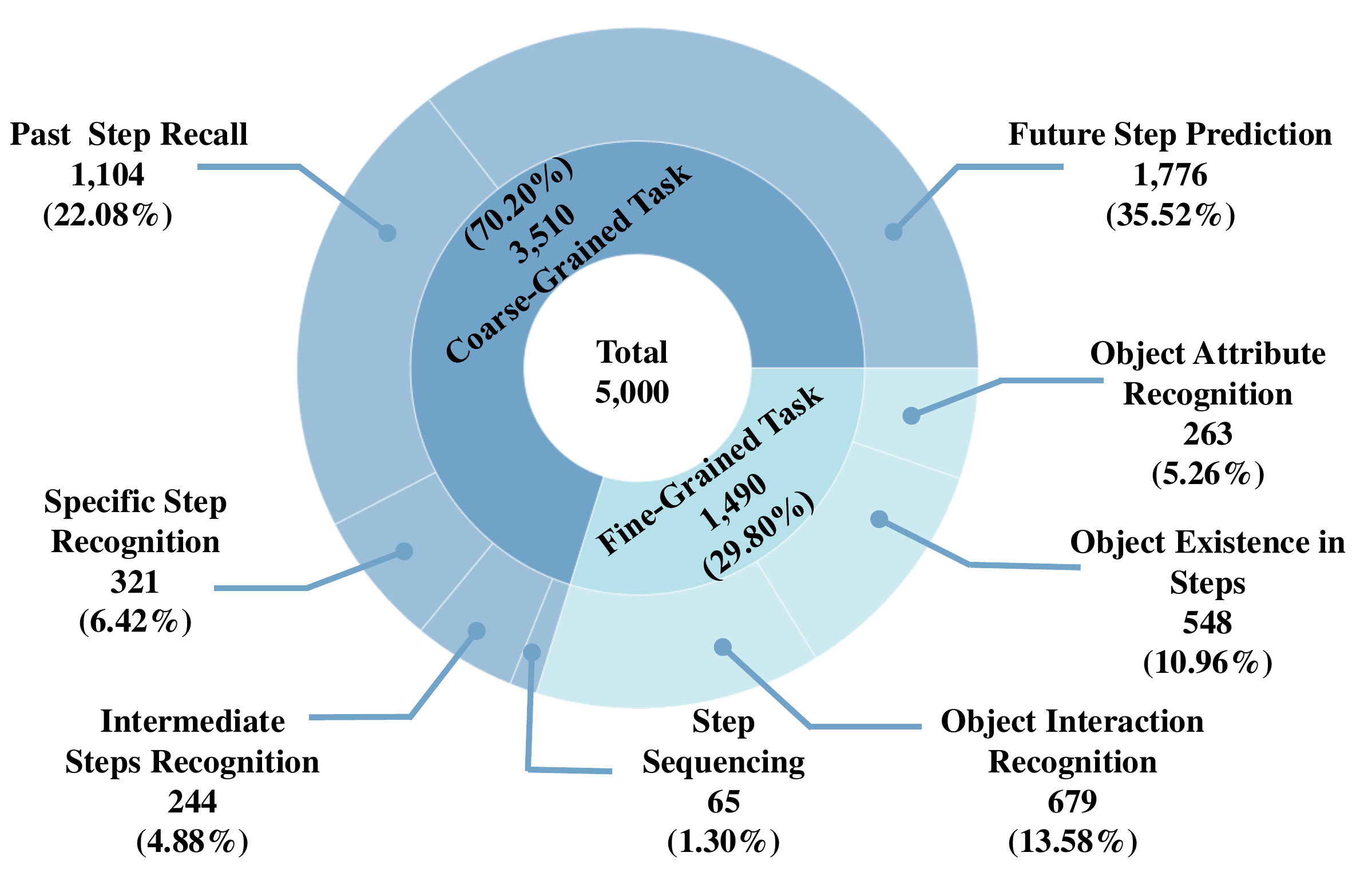}
    \end{minipage}
    \caption{Overview of InstructionBench. Left: Detailed statistics of InstructionBench. Right: Task type distribution of InstructionBench, focusing on evaluating Coarse-Grained (event-level) and Fine-Grained (object-level) temporal reasoning ability in instructional video scenarios.}
    \label{fig:data_statistics_and_task_type_distribution}
\end{figure*}

\section{InstructionBench}


In this section, we demonstrate the process of constructing our InstructionBench and training dataset. Figure~\ref{fig:data_construction_process} shows the overall construction process, Figure~\ref{fig:example_case1} and Table~\ref{tab:new_task_dimension} show examples from our InstructionBench.



\subsection{Dataset Collection}
To ensure the ultimate quality of our InstructionBench, we meticulously select data sources that encompass a broad range of instructional scenarios and feature high-quality annotations, specifically those manually annotated. 

Firstly, we select YouCook2~\cite{zhou2018towards} and HiREST~\cite{zala2023hierarchicalvideomomentretrievalstepcaptioning} for their third-person view instructional videos, which surpass similar datasets ~\cite{tang2019coin, zhukov2019cross} due to their high-quality, manually curated annotations. YouCook2 specializes in cooking, and HiREST expands on this with a wider range of scenarios like home, garden, and vehicle maintenance. As interest in first-person view~(e.g. egocentric) video research grows, we add egocentric videos to our collection, including Ego4D Goal-Step~\cite{song2024ego4d}, and Ego-Exo4D~\cite{grauman2024ego} for a comprehensive view. The former caters specifically to cooking scenarios, while the latter covers a broad range of human skill activities, from cooking to bike repair and health-related tasks.  The statistics of these datasets are show in Table~\ref{tab:source_datasets_data_statistics}. 

Notably, Ego4D Goal-Step videos have an average duration of 26 minutes, far surpassing other collection lengths, both Ego4D Goal-Step and Ego-Exo4D show step repetition. Therefore, we implement trimming to standardize lengths and reduce the repetition rates of steps (See the Appendix for more details). We create questions based on the step annotations of videos from the validation split of each dataset to establish our InstructionBench. For YouCook2 and HiREST datasets, we utilize it original datasets. For Ego4D Goal-Step and Ego-Exo4D, we employ their trimmed versions as stated before.

\begin{table}[!t]
    \centering
    \scalebox{0.8}{
    \tabcolsep=2pt
    \begin{tabular}{lcccccc}
    \toprule
        Dataset & View & Len.(mins) & \#Avg.Steps & \#Train & \#Val \\ 
    \midrule
    YouCook2* & Third & 5.26 & 7.72 & 1,333 & 457 \\ 
    HiREST* & Third & 4.44 & 7.56 & 550 & 78 \\ 
    Ego4D Goal-Step & First & 26.01 & 23.28 & 583 & 134 \\ 
    Ego-Exo4D & First & 5.03 & 26.68 & 686 & 182 \\ 
    Ego4D Goal-Step* (trimmed) & First & 4.98 & 11.48 & 1,045 & 253 \\ 
    Ego-Exo4D* (trimmed) & First & 4.29 & 18.58 & 473 & 158 \\ 
    \bottomrule
    \end{tabular}
    }
    \caption{\small Data statistics of source datasets for InstructionBench. ``Third'' means videos in this dataset are third-person view, ``First'' means first-person view. Len.(mins): The average length of the videos in minutes. \#Avg.Steps: The average number of steps. \#Val/\#Train: The number of training/validation videos. We use the dataset marked with *, where the validation part is used for InstructionBench, and the training part for the Instructional Video Training Dataset. 
    }
    \label{tab:source_datasets_data_statistics}
\end{table}

\subsection{Automated Open-Ended Q\&A Generation}
In the data construction phase, we leverage GPT-4 to generate Q\&As using the step annotations from the original datasets, which include descriptions and timestamps for actions/events occurring in instructional videos.

\subsubsection{Design of Temporal Reasoning Prompts.}
We design prompts that included step annotations and under the informed guidance of instructional videos' characteristics to direct GPT-4 in generating Q\&A for temporal reasoning. We structure prompts divided into two main categories: Coarse-Grained and Fine-Grained.
\begin{itemize}
    \item The \textbf{Coarse-Grained} concentrates on the event level, directing GPT-4 to formulate questions about action sequences and identify specific actions. To correctly answer these questions, Video-LLMs are asked to identify and sequence key actions using temporal reasoning to comprehend the timing of each event in the video.
    \item The \textbf{Fine-Grained} instructs GPT-4 to inquire about objects appearing in the video, but the ultimate questions still relate to specific actions, as our focus is on understanding the actions occurring in the video, rather than static visual analysis. Answering correctly requires Video-LLMs to use temporal reasoning to not only identify these objects but also to determine their sequence and relationship to the action timeline in the video.
\end{itemize}
We also tailor questions to match the dataset perspective. For example, for the first-person datasets like Ego4D Goal-Step and Ego-Exo4D, we instruct GPT-4 to create questions from the first-person viewpoint of a user who interacts with a smart eyeglass assistant, designed to help recall past events and predict future activities. Additionally, we prompt GPT-4 to use varied vocabulary to naturally reference steps and actions, designing diverse and challenging questions with concise and brief answers, each under 20 words. Detailed prompts are provided in the Appendix.



\subsubsection{Setting of Q\&A Quantity.}
The average number of step annotations per dataset determines the number of Q\&As. For Ego4D and Ego-Exo4D, with 11-18 steps, we set 20 Q\&As per video. For HiREST and YouCook, with about 7 steps, we set 10 Q\&As per video to prevent repetitive Q\&As.

\subsubsection{Question Categorization.}
After establishing a comprehensive outline for question generation, we combine manual efforts with GPT-4's assistance to categorize questions. Initially, we manually check random GPT-4-generated Q\&As to create a preliminary set of categories. Using these, GPT-4 then classifies the remaining Q\&As. Questions that didn't fit are labeled as `Other'. We conclude with a manual review, randomly checking GPT-4's classifications and re-examining `Other' questions. This iterative process ensure reliable categorization. For question types and examples, please see Table~\ref{tab:new_task_dimension}.

\subsection{Quality Assurance for Generated Q\&A Pairs}

After manually reviewing the Q\&As generated by GPT-4, we identify several issues such as questions being answerable without watching the video or being too ambiguous. To address this, we introduce filtering strategies to ensure benchmark quality. All GPT-4-generated Q\&As undergo a filtering chain by multiple AI assistants, retaining only those that pass all checks. Before detailing the process, we introduce the notations used: video's original step annotation ($\mathcal{V}$), Question ($\mathcal{Q}$), and Answer ($\mathcal{A}$). The filtering chain consists of several stages, each utilizing different combinations of this information. Key stages include:

\begin{itemize}
    \item \textbf{Blind question filtering.} Firstly, we eliminate questions that could be correctly answered without watching the video content, primarily involving common-sense knowledge, which is often found in instructional videos. Without this filtration, multimodal models might rely solely on their language components, hindering visual development. To achieve this, we reference Egoschema~\cite{mangalam2023egoschemadiagnosticbenchmarklongform} and make it stricter by using more LLM assistants to reduce missed detections. Specifically, AI assistants (GPT-4o, Gemini Pro, and Claude 2) respond based solely on $\mathcal{Q}$. If any provided a correct answer, the question was categorized as a blind question. This rigorous method eliminate 11\% to 31\% of the original data, significantly highlighting that our benchmark focuses on visual elements rather than common-sense understanding.

    \item \textbf{Hallucination question filtering.} GPT-4 occasionally generates questions unrelated to the video's step annotations, making them ``unanswerable.'' Since GPT-4 only has access to the step annotations, questions beyond these annotations introduce hallucinations. Therefore, we use three AI assistants to evaluate the relevance of $\mathcal{Q}$ and $\mathcal{V}$. If the majority (over half) deem the question unrelated, that Q\&A pair is removed. This process eliminates 3\% to 31\% of the original data.

    \item \textbf{Tricky question filtering.} GPT-4 sometimes generates excessively complex questions that remain unanswerable even after thorough video review. To ensure questions are manageable and answerable, we eliminate tricky Q\&As. Specifically, three AI assistants respond based on $\mathcal{V}$ and $\mathcal{Q}$. If all three fail to provide correct answers, the Q\&A is deemed tricky and removed. This process filters out 6\% to 14\% of the original data.
    
    \item \textbf{Unspecific and incomplete answers filtering.} GPT-4's designed $\mathcal{A}$ can sometimes be incomplete or lack specificity. To address this, we use three AI assistants to evaluate and remove such answers based on $\mathcal{V}$, $\mathcal{Q}$, and $\mathcal{A}$. This step filters out 11\% to 37\% of the original data.
\end{itemize}

\subsection{Multiple-Choice Transformation}

To facilitate clearer and simpler model evaluation on our InstructionBench, we convert the generated and filtered open-ended Q\&A pairs into a multiple-choice format. Using GPT-4, we generate incorrect answers based on the video's step annotation, question, and answer. The process involves:

\begin{itemize}
    \item  \textbf{Avoiding Duplicate Wrong Answers:} GPT-4 generated more wrong answers than needed, from which the most suitable were selected to ensure no duplicates.
    \item \textbf{Ensuring Incorrectness:} Three AI assistants checked the answers to confirm they were indeed incorrect.
    \item \textbf{Enhancing Distractors:} AI assistants assessed the complexity of distractors to ensure they were effective.
\end{itemize}
For more detailed prompts and the full transformation process, please see Appendix.

\begin{table}[t!]
    \centering
    \scalebox{0.75}{
        \tabcolsep=0pt
        \begin{tabular}{lccccc}
            \toprule
            Benchmarks & Video Source & \#Video & Len.(s) & Q\&A Format & \#Q\&A \\
            \midrule
            MSRVTT-QA & MSRVTT & 2,990 & 15.2 & OE & 72,821 \\
            MSVD-QA & MSVD & 504 & 9.8 & OE & 13,157 \\
            TGIF-QA & TGIF & 9,575 & 3.0 & MC/OE & 8,506 \\
            ActivityNet-QA & ActivityNet & 800 & 111.4 & OE & 8,000 \\
            TVQA & TV show & 2,179 & 11.2 & MC & 15,253 \\
            How2QA & HowTo100M/TV & 1,166 & 15.3 & MC & 2,852 \\
            STAR & Charades & 914 & 11.9 & MC & 7,098 \\
            NExT-QA & YFCC-100M & 1,000 & 39.5 & MC/OE & 8,564 \\
            \midrule
            VideoBench & Combined & 5,917 & 56.0 & MC & 17,036 \\
            MVBench & Combined & 3,641 & 16.0 & MC & 4,000 \\
            EgoSchema & Ego4D & 5,063 & 180.0 & MC & 5,063 \\
            VideoMME & YouTube & 900 & 1,017.9 & MC & 2,700 \\
            TempCompass & ShutterStock & 410 & 11.4 & MC/OE & 7,540 \\
            VideoVista & YouTube & 3,402 & 131.0 & MC/OE & 24,906 \\
            AutoEval-Video & YouTube & 327 & 14.6 & OE & 327 \\
            LVBench & YouTube & 103 & 4,101.0 & MC & 1,549 \\
            \midrule
            InstructionBench(Ours) & Instructional Datasets & 713 & 282.9 & MC & 5,000 \\
            \bottomrule
        \end{tabular}
        }
    \caption{\small The comparison of various benchmarks. \#Videos: the total number of videos. Len.(s): the average length of the videos in seconds. For Q\&A format, MC means Multiple Choice, OE means Open-Ended. \#Q\&A: The number of Q\&A pairs.}
    \label{tab:comparison_benchmark}
\end{table}

\subsection{InstructionBench Statistics} 
After the above data construction and filtering process, finally, as shown in Figure~\ref{fig:data_statistics_and_task_type_distribution}, our InstructionBench contains 932 video clips merged from multiple instructional video datasets, with an average duration of 282.95 seconds. Our benchmark targets temporal reasoning tasks at two granularities, with the task distribution shown in the right subplot of Figure~\ref{fig:data_statistics_and_task_type_distribution}.


\begin{table*}[t!]
    \centering
    \scalebox{0.9}{
    \tabcolsep=2.5pt
    \begin{tabular}{lcccccccccccccc} 
        \toprule
        \multirow{2}{*}{Model} & \multirow{2}{*}{\# Frames} & \multicolumn{5}{c}{Coarse-Grained Task(\%)} & \multicolumn{3}{c}{Fine-Grained Task(\%)} & \multirow{2}{*}{CG Overall(\%)} & \multirow{2}{*}{FG Overall(\%)} & \multirow{2}{*}{Overall(\%)} \\ 
        \cmidrule(lr){3-7} \cmidrule(lr){8-10}
         & & FSP & PSR & ISR & SS & SSR & OES & OAR & OIR & & & \\ 
        \midrule
        \textit{Random} & - & 18.81 & 22.37 & 17.21 & 18.46 & 18.69 & 20.26 & 21.29 & 19.59 & 19.80 & 20.13 & 19.90 \\
        \midrule
        Video-LLaVA & 8 & 23.54 & 26.27 & 20.49 & 27.69 & 32.71 & 25.73 & 29.28 & 28.42 & 25.10 & 27.58 & 25.84 \\
        LLaVA-NeXT-Video & 8 & 24.16 & 24.00 & 24.59 & 29.23 & 29.91 & 27.74 & 29.28 & 30.49 & 24.76 & 29.26 & 26.10 \\
        VideoChat2 & 8 & 27.59 & 26.99 & 33.20 & 36.92 & 43.93 & 33.39 & 39.54 & 37.41 & 29.46 & 36.31 & 31.50 \\
        VideoLLaMA2 & 8 & 26.80 & 28.53 & 31.56 & 30.77 & 40.19 & 37.23 & 36.88 & 38.44 & 28.97 & 37.72 & 31.58 \\
        
        \midrule
        LLaVA-NeXT-Video & 16 & 23.76 & 23.64 & 22.95 & 27.69 & 26.48 & 26.09 & 26.62 & 25.77 & 23.99 & 26.04 & 24.60 \\
        VideoChat2 & 16 & 27.93 & 27.54 & 34.43 & 36.92 & 42.37 & 36.50 & 39.16 & 40.94 & 29.74 & 38.99 & 32.50 \\
        VideoLLaMA2 & 16 & 26.13 & 27.08 & 34.43 & 27.69 & 39.88 & 36.13 & 41.83 & 33.28 & 28.29 & 35.84 & 30.54 \\
        \midrule
        LLaMA-VID & 1 fps & 23.09 & 21.92 & 20.08 & 33.85 & 33.64 & 24.45 & 23.95 & 24.89 & 23.68 & 24.56 & 23.94 \\
        \midrule
        Gemini Pro Vision & 8 & 37.33 & 39.40 & 35.66 & 33.85 & 53.27 & 47.63 & 49.05 & 50.96 & 39.26 & 49.40 & 42.28 \\
        GPT-4V & 8 & 37.73 & 39.49 & 49.18 & 38.46 & 49.53 & 43.43 & 49.05 & 46.39 & 40.17 & 45.77 & 41.84 \\
        GPT-4o & 8 & 44.71 & 43.39 & 55.33 & 43.08 & 59.50 & 47.26 & 55.89 & 58.91 & 46.35 & 54.09 & 48.66 \\
        \midrule
        Gemini Pro Vision & 16 & 38.85 & 41.03 & 47.95 & 30.77 & 53.27 & 47.99 & 52.09 & 52.28 & 41.34 & 50.67 & 44.12 \\
        GPT-4V & 16 & 41.55 & 40.40 & 55.74 & 40.00 & 52.96 & 48.54 & 50.19 & 53.02 & 43.19 & 50.87 & 45.48 \\
        GPT-4o & 16 & 49.49 & 48.91 & 57.79 & 40.00 & 62.31 & 54.74 & 62.36 & 62.00 & 50.88 & 59.40 & 53.42 \\
        \bottomrule
    \end{tabular}
    }
    \caption{Comprehensive evaluation results of various Video-LLMs on InstructionBench with different frame settings. FSP: Future Step Prediction, PSR: Past Step Recall, ISR: Intermediate Steps Recognition, SS: Step Sequencing, SSR: Specific Step Recognition; OES: Object Existence in Steps, OAR: Object Attribute Recognition, OIR: Object Interaction Recognition; CG Overall: Coarse-Grained Overall, FG Overall: Fine-Grained Overall.} 
    \label{tab:evaluation_results_2}
\end{table*}

We also compare our benchmark to others in Table~\ref{tab:comparison_benchmark}. Early benchmarks (from MSRVTT-QA to Next-QA) featured shorter or limited-domain videos. Recent benchmarks aimed for comprehensive video understanding and reasoning for the latest Video-LLMs. Distinct from them, we focus on temporal reasoning in instructional videos. 

\subsection{Instructional Video Training Dataset Collection}
Based on the automated framework for generating high-quality Q\&A pairs, which constructs InstructionBench, we also create a training set from the training splits of source datasets for the fine-tuning stage of Video-LLMs. This set comprises over 19k Q\&A pairs in total. We present the training Q\&As in an open-ended format, which aligns with the widely used fine-tuning data format. The statistics for the training set are provided in Appendix.

\section{Experiments}
In this section, we conduct an evaluation of prevalent models on our InstructionBench, covering both open-source models like Video-LLaVA~\cite{lin2023video}, LLaVA-NeXT-Video~\cite{zhang2024llavanextvideo},  VideoChat2~\cite{li2024mvbench}, VideoLLaMA2~\cite{cheng2024videollama} and LLaMA-VID~\cite{li2023llama}, as well as closed-source ones including GPT-4o, GPT-4V, Gemini Pro Vision.  For a more direct comparison, we report the accuracy of each model on multiple-choice questions in our InstructionBench. Apart from LLaMA-VID and Video-LLaVA which have set frame rate specifications, we uniformly sample different video frame counts (\textit{e.g.}, 8/16 frames) for the same model to assess the impact of input frame counts on model performance.

\subsection{Overall Results}

As illustrated in Table~\ref{tab:evaluation_results_2}, closed-source models comprehensively outperform open-source models, highlighting a substantial gap. Among open-source Video-LLMs, VideoLLaMA2 and VideoChat2 show better performance due to extensive training data and the incorporation of both temporal and spatial modeling. VideoChat2 uses UMT~\cite{li2023unmasked} to enhance temporal and spatial learning, while VideoLLaMA2 employs a Space-Time Convolution Connector (STC)~\cite{cheng2024videollama} for balanced feature capture. In contrast, LLaMA-VID and LLaVA-NeXT-Video face challenges due to limited training data and lack of specialized temporal and spatial modeling. Considering that our InstructionBench contains many questions requiring temporal reasoning, the ability of Video-LLMs to capture spatio-temporal correlations is critical for good performance.

Increasing the number of input frames significantly enhances the performance of closed-source models. GPT-4o shows a 4.76\% improvement when we input 16 frames instead of 8 frames to it. GPT-4V and Gemini Pro Vision also get better results when the input video frames increase. However, for open-source models, increasing frame input offers little to no benefit and can even decrease performance, especially for those lacking temporal and spatial modeling. For example, LLaVA-NeXT-Video sees a 1.5\% overall performance drop in our InstructionBench when the input video frames increase from 8 to 16. Despite processing more frames at 1 frame per second, LLaMA-VID underperforms compared to 8-frame Video-LLaVA, which uses LanguageBind's~\cite{zhu2023languagebind} video encoder for better temporal and spatial attention integration. Furthermore, LLaMA-VID compresses one video frame to just 1-2 tokens, resulting in significant information loss of video frames. This approach leads to LLaMA-VID achieving the lowest results in Fine-Grained tasks on our InstructionBench. These findings suggest that existing Video-LLMs must improve their ability to process multiple frames to fully capture video information.

\begin{figure*}[t!]
    \centering
    \begin{minipage}[t]{0.49\textwidth}
        \centering
        \includegraphics[width=\textwidth]{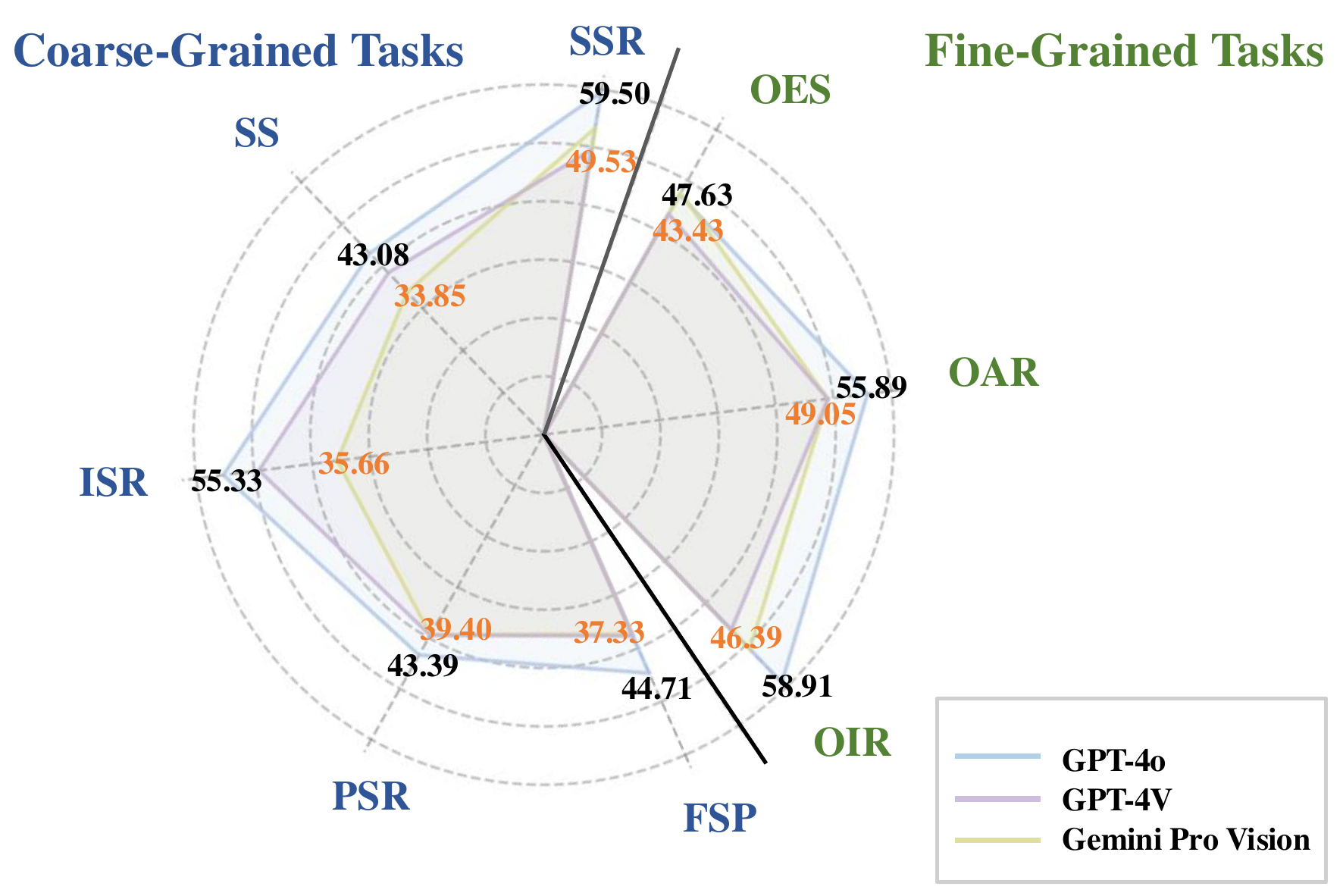}
    \end{minipage}
    \begin{minipage}[t]{0.49\textwidth}
        \centering
        \includegraphics[width=\textwidth]{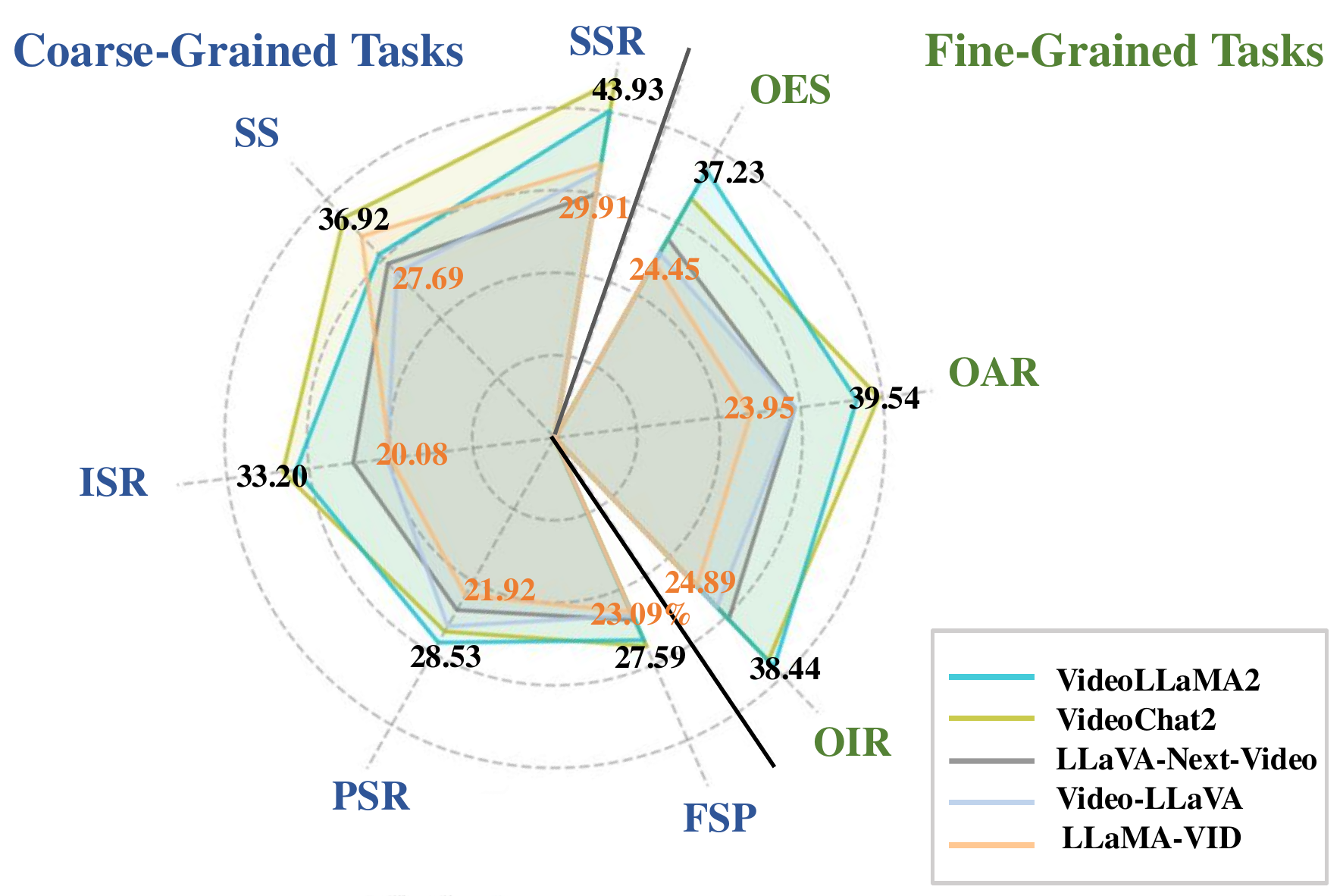}
    \end{minipage}
    \caption{Comparison of Closed-Source and Open-Source Models' Evaluation Results divided by Task Granularity. We only mark the highest (with black color) and lowest (with orange color) scores for each task in the figure. The abbreviations used in the figure are as follows: SSR: Specific Step Recognition, SS: Step Sequencing, ISR: Intermediate Steps Recognition, PSR: Past Step Recall, FSP: Future Step Prediction, OES: Object Existence in Steps, OAR: Object Attribute Recognition, OIR: Object Interaction Recognition.}
\label{fig:comparison_models_evaluation_results_by_granularity}
\end{figure*}

\subsection{Detailed Results by Task Granularity}
We continue to evaluate the performance of each model across tasks of varying granularity. For a fair comparison, we use an 8-frame input setting for all models, with the exception of LLaMA-VID. Furthermore, due to the significant gap between open source and closed source, we have made a separate visual comparison for the open-source and closed-source groups, as shown in Figure ~\ref{fig:comparison_models_evaluation_results_by_granularity}. Detail numerical results of these models are shown in Table~\ref{tab:evaluation_results_2}.


For the closed-source models, GPT-4o maintains a significant edge in Coarse-Grained tasks. Between GPT-4V and Gemini Pro Vision, neither shows a distinct lead overall. However, GPT-4V excels in recognizing sequences, such as Past Step Recall, Future Step Prediction, Intermediate Steps Recognition, and Step Sequencing, showcasing its robust temporal reasoning ability. Conversely, Gemini Pro Vision performs better in Specific Step Recognition, requiring precise identification and localization of particular steps. This skill underscores its strength in temporal reasoning by understanding the context of each step. Additionally, Gemini Pro Vision surpasses GPT-4V in Fine-Grained tasks and outperforms GPT-4o in object Existence in Steps, highlighting its superior detail-capturing capability.

For the open-source models, we can observe that VideoChat2 and VideoLLaMA2 demonstrate significant advantages in tasks both Coarse-Grained and Fine-Grained. Meanwhile, although LLaMA-VID generally lags in comprehensive results, it outperforms VideoLLaMA2 in the Step Sequencing task and excels over LLaVA-NeXT-Video in Specific Step Recognition. These results suggest that LLaMA-VID gains a moderate advantage through its input processing at 1 frame per second, thus affirming the efficacy of its dual-token representation in handling long-duration video sequences.

\subsection{Instructional Video Training Dataset Value}



\begin{table}[!t]
    \centering
    \scalebox{0.72}{
    \begin{tabular}{lccccc} 
        \toprule
        Model & \# Frames & CG Overall(\%) & FG Overall(\%) & Overall(\%) \\ 
        \midrule
        LLaMA-VID & 1 fps & 23.68 & 24.56 & 23.94 \\
        LLaMA-VID$\boldsymbol{+}$ & 1 fps & 32.17 & 34.70 & 32.92 \\
        \midrule
        Video-LLaVA & 8 & 25.10 & 27.58 & 25.84 \\
        Video-LLaVA$\boldsymbol{+}$ & 8 & 26.30 & 27.25 & 26.58 \\
        \bottomrule
    \end{tabular}
    }
    \caption{\small The `` $\boldsymbol{+}$'' indicates a model whose supervised fine-tuning (SFT) stage also incorporates our instructional video training dataset. CG: Coarse-Grained Overall, FG: Fine-Grained Overall.}
\label{tab:after_finetune_on_instrutional_training_data_evaluation_results}
\end{table}

We incorporate our created instructional video training dataset to the supervised fine-tuning (SFT) stage of LLaMA-VID and Video-LLaVA. Both models have less training data compared to VideoLLaMA2 and VideoChat2, allowing us to preliminarily evaluate the effectiveness of our training set. The results are shown in Table~\ref{tab:after_finetune_on_instrutional_training_data_evaluation_results}. The models trained with the addition of our instructional dataset to their original SFT datasets are referred to as LLaMA-VID$\boldsymbol{+}$ and Video-LLaVA$\boldsymbol{+}$. The most substantial improvement is observed in LLaMA-VID, with its overall performance increasing to 32.92\% from the initial 23.94\%. Such results demonstrate the effectiveness of our created training dataset to improve Video-LLM's ability in understanding instructional scenarios.

Video-LLaVA shows a modest performance increase from 25.84\% to 26.58\%. This enhancement is attributed to its smaller number of trainable parameters compared to LLaMA-VID. Unlike LLaMA-VID's Q-former tuning, Video-LLaVA only tunes a share projection layer during fine-tuning, resulting in a relative shortfall in adaptive learning on the new dataset.

\section{Conclusion}


In this paper, we present InstructionBench, a benchmark designed to evaluate the temporal reasoning capabilities of Video-LLMs in instructional video scenarios. We create an automated framework for generating Q\&A pairs and compile a comprehensive instructional video dataset, providing valuable resources for the field. Our evaluation of popular Video-LLMs shows that closed-source models like GPT-4o significantly outperform open-source ones, highlighting a gap in temporal reasoning performance. Additionally, our findings indicate that current models struggle with fine-grained temporal understanding, pointing to a need for further improvements.

\newpage
\bibliography{references}

\end{document}


\maketitle

\section{Abstract}
This supplementary material includes the following contents:
\begin{itemize}
    \item Data analytics for our created InstructionBench as well as our proposed Instructional Video Training dataset, which encompasses the distribution of correct answer option symbols within our benchmark, the distribution of video length, and other detailed statistics.
    \item Evaluation details, which illustrate the process of how we evaluate different video large language models (Video-LLMs) in our proposed InstructionBench.
    \item Video trimming strategy for Ego4D Goal-Step and Ego-Exo4D, which details the procedure followed to trim videos within these specific datasets.
    \item Multiple-Choice Transformation, which describes the process of how we transform open-ended Q\&A pairs into multiple-choice format.
    \item A prompt library that lists the prompt templates used in our automated Q\&A generation framework.
\end{itemize}

\section{A. Additional Data Analytics}
\subsection{A1. Correct Answer Option Symbols Distribution}
\begin{figure}[!h]
    \centering
    \includegraphics[width=.3745\textwidth]{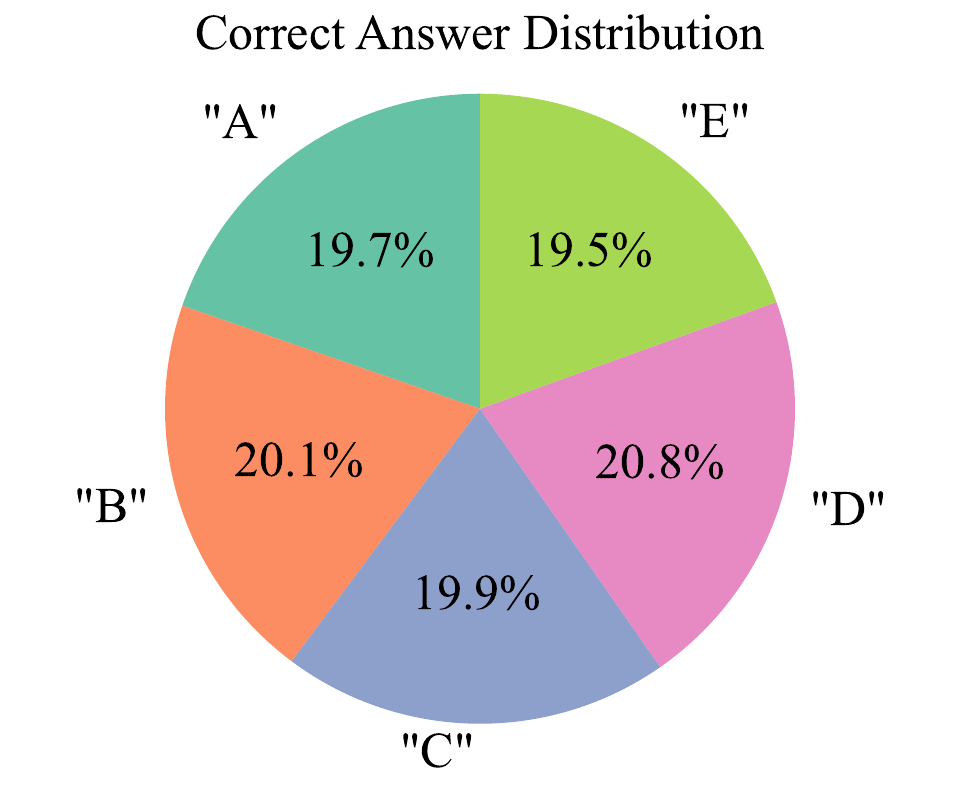}
    \caption{Correct Answer Option Symbols Distribution in InstructionBench.}
    \label{fig:answer_distribution_pie_chart}
\end{figure}

To prevent models from simply guessing the correct answers, we have shuffled the options, ensuring that the correct answers are evenly distributed across different option symbols. The Figure~\ref{fig:answer_distribution_pie_chart} illustrates the distribution of these correct answer option symbols within our dataset, revealing a uniform distribution that reflects the deliberate design and fairness integral to our InstructionBench.

\subsection{A2. Video Length Distribution}
\begin{figure*}[!t]
    \centering
    \begin{minipage}[t]{0.45\textwidth}
        \centering
        \includegraphics[width=\textwidth]{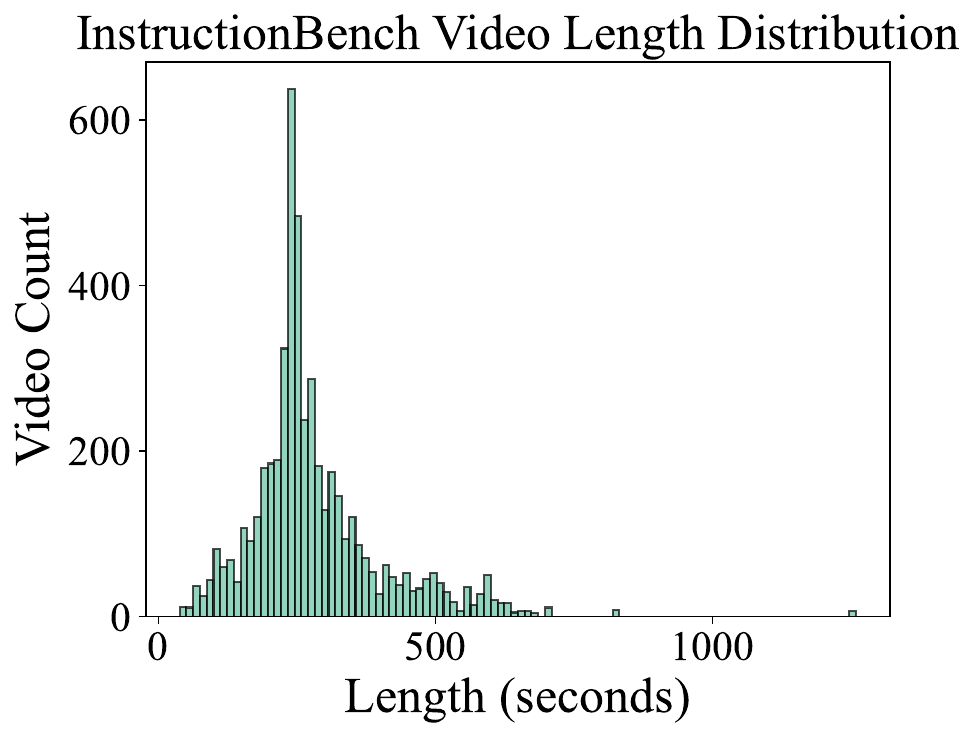}
        \caption{Video Length Distribution in InstructionBench.}
        \label{fig:instructionbench_video_length_distribution}
    \end{minipage}
    \hfill
    \begin{minipage}[t]{0.45\textwidth}
        \centering
        \includegraphics[width=\textwidth]{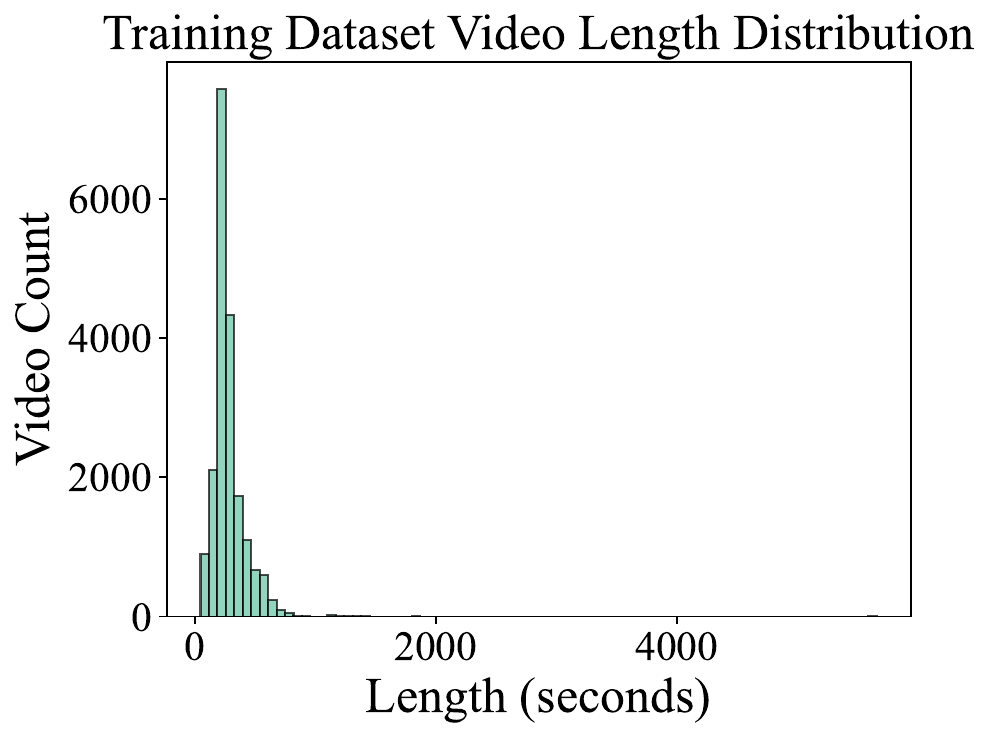}
        \caption{Video Length Distribution in our created Instructional Video Training Dataset.}
        \label{fig:training_dataset_video_length_distribution}
    \end{minipage}
\end{figure*}

We present the video duration distributions in our InstructionBench and our created Instructinal Video Training Dataset in Figure~\ref{fig:instructionbench_video_length_distribution} and Figure~\ref{fig:training_dataset_video_length_distribution}, respectively.

\subsection{A3. Data Statistics of InstructionBench and Instructional Video Training Dataset}
Detailed Statistics of InstructionBench and Instructional Video Training Dataset are shown in Table~\ref{tab:instructionbench_data_mcq_format_statistics_appendix}, Table~\ref{tab:instructionbench_data_open_ended_format_statistics_appendix} and Table~\ref{tab:training_dataset_data_statistics_appendix}.

Our InstructionBench features two formats: multiple-choice and open-ended questions, with their respective statistics detailed in Table~\ref{tab:instructionbench_data_mcq_format_statistics_appendix} and Table~\ref{tab:instructionbench_data_open_ended_format_statistics_appendix}. Following the process of converting open-ended Q\&As to multiple-choice format, as discussed in the Multiple-Choice Transformation section, if GPT-4 cannot successfully transform an open-ended Q\&A within 10 attempts, we discard that particular Q\&A from the conversion process.
This results in a discrepancy in the number of Q\&As between the two formats.

For the Instructional Video Training Dataset, we exclusively creat Q\&As with open-ended format, which aligns with the widely used fine-tuning data format. The specifics of these statistics are shown in Table~\ref{tab:training_dataset_data_statistics_appendix}.

\section{B. Evaluation Details}
In this section, we will demonstrate how we evaluate different Video-LLMs in our proposed InstructionBench, covering both open-source models like Video-LLaVA~\cite{lin2023video}, LLaVA-NeXT-Video~\cite{zhang2024llavanextvideo},  VideoChat2~\cite{li2024mvbench}, VideoLLaMA2~\cite{cheng2024videollama} and LLaMA-VID~\cite{li2023llama}, as well as closed-source ones including GPT-4o~\cite{openai2024gpt4o}, GPT-4V~\cite{gpt4v}, Gemini Pro Vision~\cite{reid2024gemini}.

\subsubsection{Video-LLaVA:} We evaluate Video-LLaVA by loading its officially-provided Video-LLaVA-7B checkpoint, and using inputs of 8 frames. During the evaluation process, we use the prompt:\\
\texttt{
Carefully watch the video and pay attention to the cause and sequence of events, the detail and movement of objects, and the action. Based on your observations, select the best option that accurately addresses the question.
<Video>
\\Question:xxx \\Options:xxx 
\\Answer with the option's letter from the given choices directly.
}

\subsubsection{LLaVA-NeXT-Video:} We evaluate by loading its officially-provided LLaVA-NeXT-Video-7B-hf checkpoint and using inputs of 8 frames and 16 frames. During the evaluation process, we use the prompt: \\
\texttt{
Carefully watch the video and pay attention to the cause and sequence of events, the detail and movement of objects, and the action. Based on your observations, select the best option that accurately addresses the question.\\
<Video>
\\Question:xxx \\Options:xxx 
\\Answer with the option's letter from the given choices directly.
}

\subsubsection{VideoChat2:} We evaluate VideoChat2 by loading its officially-provided videochat2\_mistral\_7b\_stage3.pth checkpoint, and using inputs of 8 frames and 16 frames. During the evaluation process, we refer to the prompt provided by MVBench~\cite{li2024mvbench}: \\\texttt{Carefully watch the video and pay attention to the cause and sequence of events, the detail and movement of objects, and the action. Based on your observations, select the best option that accurately addresses the question. \\
<Video>
\\Question:xxx \\Options:xxx 
\\Only give the best option. Best option:(}

\subsubsection{VideoLLaMA2:}
We evaluate VideoLLaMA2 by loading its officially-provided VideoLLaMA2-7B checkpoint for processing inputs of 8 frames, and the VideoLLaMA2-7B-16F checkpoint for inputs of 16 frames. During the evaluation process, we use the prompt:\\
\texttt{
Carefully watch the video and pay attention to the cause and sequence of events, the detail and movement of objects, and the action. Based on your observations, select the best option that accurately addresses the question.
<Video>
\\Question:xxx \\Options:xxx 
\\Answer with the option's letter from the given choices directly and only give the best option.
}

\begin{table*}[!t]
    \centering
    \begin{minipage}[t]{0.45\textwidth}
        \centering
        \setlength{\tabcolsep}{0.9pt}
        \renewcommand{\arraystretch}{1.2}
        \small
        \begin{tabular}{lc}
            \toprule
            \textbf{Category} & \textbf{Size} \\
            \midrule
            Video Source Datasets: (YouCook2, HiREST, \\
            Ego4D Goal-Step, Ego-Exo4D) & 4 \\
            Videos & 713 \\
            Video Clips & 932 \\
            Video Average Duration & 282.95s \\
            Average Question Length & 11.66 \\
            Average Option Length & 10.92 \\
            Option Numbers & 5 \\
            Total Samples & 5,000 \\
            \bottomrule
        \end{tabular}
        \caption{Detailed statistics of our InstructionBench with Multiple-Choice format.}
        \label{tab:instructionbench_data_mcq_format_statistics_appendix}
    \end{minipage}
    \hfill
    \begin{minipage}[t]{0.45\textwidth}
        \centering
        \setlength{\tabcolsep}{0.9pt}
        \renewcommand{\arraystretch}{1.2}
        \small
        \begin{tabular}{lc}
            \toprule
            \textbf{Category} & \textbf{Size} \\
            \midrule
            Video Source Datasets: (YouCook2, HiREST, \\
            Ego4D Goal-Step, Ego-Exo4D) & 4 \\
            Videos & 718 \\
            Video Clips & 938 \\
            Video Average Duration & 283.22s \\
            Average Question Length & 11.70 \\
            Average Answer Length & 11.01 \\
            Total Samples & 5,635 \\
            \bottomrule
        \end{tabular}
        \caption{Detailed statistics of our InstructionBench with Open-Ended format.}
        \label{tab:instructionbench_data_open_ended_format_statistics_appendix}
    \end{minipage}
\end{table*}
\begin{table}[!t]
    \centering
    \setlength{\tabcolsep}{0.9pt}
    \small
    \begin{tabular}{lc}
        \toprule
        \textbf{Category} & \textbf{Size} \\
        \midrule
        Video Source Datasets: (YouCook2, HiREST, \\
        Ego4D Goal-Step, Ego-Exo4D) & 4 \\
        Videos & 2475 \\
        Video Clips & 3346 \\
        Video Average Duration & 284.10s \\
        Average Question Length & 11.69 \\
        Average Answer Length & 10.85 \\
        Total Samples & 19,495 \\
        \bottomrule
    \end{tabular}
    \caption{Detailed statistics of Instructional Video TrainingDataset. Our Instructional Video Training Dataset only includes open-ended Q\&As.} 
    \label{tab:training_dataset_data_statistics_appendix}
\end{table}
\subsubsection{LLaMA-VID:} We evaluate LLaMA-VID by loading officially-provided llama-vid-7b-full-224-video-fps-1 checkpoint and employing a sampling rate of 1 frame per second as input. During the evaluation process, we use the prompt: \\
\texttt{
Carefully watch the video and pay attention to the cause and sequence of events, the detail and movement of objects, and the action. Based on your observations, select the best option that accurately addresses the question.\\
<Video>
\\Question:xxx \\Options:xxx 
\\Answer with the option's letter from the given choices directly.
}
\subsubsection{GPT-4o, GPT-4V, Gemini Pro Vision:} We evaluate closed-source models through official API calls and using inputs of 8 frames and 16 frames. We first uniformly sample fixed frames from the video. To avoid exceeding the upload limit, we compress the images to one-quarter of their original size and then convert them to base64 encoding for upload. During the evaluation process, we use the following prompt:\\
\texttt{
You are an advanced AI model specialized in analyzing video content. You will be given a series of frames extracted from a video. Your task is to carefully watch these frames, paying close attention to the sequence of events, object details. Based on your observations, answer the subsequent questions accurately. \\
Given Frames:<Video>
\\Question:xxx \\Options:xxx 
\\Answer with the option's letter from the given choices directly.
}

\section{C. Video Trimming Strategy}
In this section, we provide a comprehensive overview of our video trimming strategy as applied to the Ego4D Goal-Step~\cite{song2024ego4d} and Ego-Exo4D datasets~\cite{grauman2024ego}.

Ego4D Goal-Step has an average video duration of 26 minutes, far surpassing the YouCook2~\cite{zhou2018towards}, HiREST~\cite{zala2023hierarchical} datasets included in our collection, and both Ego4D Goal-Step and Ego-Exo4D show step repetition. Therefore, we implement trimming to standardize lengths and reduce the repetition rates of steps. Specifically, for each video in Ego4D Goal-Step and Ego-Exo4D, we gradually accumulate steps into a video clip, starting from the first step annotation, and then follow these rules:

\begin{enumerate}
    \item \textbf{Video Clip Duration Constraint}: Continue adding steps until the duration of the current clip exceeds 4 minutes.
    \item \textbf{Step Count Check}: Ensure that the total number of steps within the clip is greater than or equal to 7 (the number is based on the average of 7.5 step annotations for HiREST~\cite{zala2023hierarchical} and YouCook2~\cite{zhou2018towards} datasets, which are among our data sources ).
    \item \textbf{Category Diversity}: The total number of different step categories should be at least 50\% of the total number of steps. This ensures diversity of steps.
   \item \textbf{Repetition Control}: The total number of steps should be at least 50\% of the maximum number of consecutive occurrences of any single step category. This controls the repetition of steps
   \item \textbf{Next Clip Initialization}: Start the next clip from the current step and repeat the above steps until all steps in the video have been examined.
\end{enumerate}

\section{D. Multiple-Choice Transformation}
In this section, we describe the process of how to transform the open-ended Q\&A pairs, formatted as \{Question, Answer\}, into multiple-choice format, denoted as \{Question, Option1, ..., Option5\}, thereby simplifying the evaluation process by enabling direct comparison of character matches in the predicted options and ensuring both efficiency and ease of assessment. For this transformation, we continued to use GPT-4, whose main task was to generate Wrong Answers($\mathcal{WA}s$) guided by the video's step annotation ($\mathcal{V}$), Question ($\mathcal{Q}$), and Answer ($\mathcal{A}$).

Nonetheless, our initial manual inspection revealed issues with the $\mathcal{WA}s$ generated by GPT-4: 1) some were irrelevant to the video content, which could undermine the assessment's challenge; 2) others were, paradoxically, correct. Consequently, we designed a multi-stage wrong answer generation process to uphold benchmark quality:
\begin{itemize}
    \item \textbf{Stage1 - Avoiding Duplicate $\bm{\mathcal{WA}s}$}: Using \{$\mathcal{V}$, $\mathcal{Q}$, $\mathcal{A}$\} for guidance, GPT-4 was prompted to generate more $\mathcal{WA}s$ than needed, specifically 6, each mirroring the correct answer to avoid noticeable differences in length, form, or format.
    Subsequently, GPT-4 chose the most suitable 4 $\mathcal{WA}s$ from the initial list, ensuring their incorrectness and relevance, and these $\mathcal{WA}s$ were checked for any duplicates, and if any were found, the generation process would start anew.
    \item \textbf{Stage2 - Guaranteeing $\bm{\mathcal{WA}s}$' Incorrectness}: Three AI assistants(GPT-4o, Gemini Pro, and Claude 2) responded based on \{$\mathcal{V}$, $\mathcal{Q}$, \textit{Options}\}. If more than half still failed even with direct access to step annotations, it raised the possibility that some wrong answers were actually correct, necessitating another round of redesign for the options.
    \item \textbf{Stage3 - Enhancing $\bm{\mathcal{WA}s}$' Distractors}: Three AI assistants responded based solely on \{$\mathcal{Q}$, \textit{Options}\}. If more than half could identify the correct answer without access to the video's step annotations, indicating insufficient complexity, then a redesign of the options was initiated.
\end{itemize}
If GPT-4 cannot successfully transform an open-ended Q\&A within 10 attempts, we discard that particular Q\&A from the conversion process.
\section{E. Prompt Library}





In this section, we provide all the prompt templates used in our automated Q\&A generation framework that unifies the processes of generating and filtering Q\&A pairs.

\subsection{E1. Q\&A Generation} 
This subsection includes prompt templates specifically designed to generate Q\&A tailored to each dataset. Each dataset will contain both Coarse-Grained and Fine-Grained templates. Refer to Table~\ref{tab:Prompt Template for Hirest Coarse-Grained Q&A Generation}, Table~\ref{tab:Prompt Template for Hirest Fine-Grained Q&A Generation}, Table~\ref{tab:Prompt Template for YouCook2 Coarse-Grained Q&A Generation}, Table~\ref{tab:Prompt Template for YouCook2 Fine-Grained Q&A Generation}, Table~\ref{tab:Prompt Template for Ego4D Goal-Step Coarse-Grained Q&A Generation}, Table~\ref{tab:Prompt Template for Ego4D Goal-Step Fine-Grained Q&A Generation}, Table~\ref{tab:Prompt Template for Ego-Exo4D Coarse-Grained Q&A Generation}, Table~\ref{tab:Prompt Template for Ego-Exo4D Fine-Grained Q&A Generation} for specific details.

Additionally, the prompt template for question categorization is also provided in Table~\ref{tab:Prompt Template for Question Categorization}.

\subsection{E2. Q\&A Filtering} 
This subsection includes prompt templates designed for the filtering stage, which encompasses Blind question filtering, Hallucination question filtering, Tricky question filtering, and Unspecific and incomplete answers filtering. Refer to Table~\ref{tab:Prompt Template for Video Question Answering Without Video and Blind Question Filtering}, Table~\ref{tab:Prompt template for Hallucination Question Filtering}, Table~\ref{tab:Prompt Template for Video Question Answering With Video and Tricky Question Filtering}, Table~\ref{tab:Prompt Template for Unspecific and Incomplete Answers Filtering} for the specific prompt templates associated with each filtering category.

In addition, the prompt template is provided to evaluate the consistency of the model's predicted answers with the correct answers, as detailed in Table~\ref{tab:Prompt Template for Checking Consistency Between Predicted and Correct Answers}.
\subsection{E3. Multiple-Choice Transformation} 
This subsection details the prompt templates involved in converting open-ended Q\&As to multiple-choice format, principally concerning the generation of suitable wrong answers based on existing open-ended Q\&As, as discussed in the Multiple-Choice Transformation section. The transformation process involves first generating six wrong answers, then selecting the four that are most suitable. The next stages include ensuring that wrong answers are incorrect and enhancing their distractors. The related prompts for these stages can be found in Table~\ref{tab:Prompt Template for 6 Wrong Answers Generation}, Table~\ref{tab:Prompt Template for Picking Up 4 Wrong Answers}, Table~\ref{tab:Prompt Template for Choosing The Best Option Without Video and Ensuring Options' Incorrectness}, Table~\ref{tab:Prompt Template for Choosing The Best Option With Video and Enhancing Options' Distractors}.

In addition, the prompt template is provided to evaluate the consistency of the model's predicted options with the correct options, as detailed in Table~\ref{tab:Prompt Template for Checking Consistency Between Predicted Option and Correct Option}.

\begin{table*}[t!]\centering
\begin{minipage}{0.95\textwidth}
\centering
\begin{tcolorbox} 
    \centering
      \small
    \begin{tabular}{p{0.95\textwidth}}
   \textcolor{blue}{\textbf{System message}} \\
You are an experienced teacher who are good at understanding long instructional video, capturing the action and movement of the video, and deeply understanding the meaning behind these movements.
Your task is to test students' ability to capture action sequence and Fine-Grained action after watching this long video. \\
    \midrule
   \textcolor{blue}{\textbf{Prompt}} \\
\#\# Instruction:\\
    \hspace{3mm}- User will give you video information in text, including the theme of the video(\texttt{"video theme"}), video's duration(\texttt{"duration"}), and key action within specific time intervals.\\
    \hspace{3mm}- Watch the video through the text video info carefully, and design three inferential, challenging question-answer pairs as if you are watching the video.\\
\#\#  Input:\\
\hspace{3mm}\textcolor{DeepBlue}{- video\_info:}\\
\hspace{6mm}\textcolor{DeepBlue}{- video theme: Make a Plaster Mask}\\
\hspace{6mm}\textcolor{DeepBlue}{- duration: 103.0s}\\
\hspace{6mm}\textcolor{DeepBlue}{- key action within specific time intervals:}\\
\hspace{9mm}\textcolor{DeepBlue}{- 35s $\sim$ 48s : clean out the face}\\
\hspace{9mm}\textcolor{DeepBlue}{- 48s $\sim$ 60s : apply tissue using water on face}\\
\hspace{9mm}\textcolor{DeepBlue}{- 60s $\sim$ 72s : apply it for full face}\\
\hspace{9mm}\textcolor{DeepBlue}{- 72s $\sim$ 77s : put it under the neck}\\
\hspace{9mm}\textcolor{DeepBlue}{- 77s $\sim$ 79s : dry it out}

\#\#  Output:\\
    \hspace{3mm}- Use your extensive experience as much as possible to design three diverse, challenging, inferential questions that fully test students' ability to capture action sequence and Fine-Grained action after watching this long video. Below are some question types you can refer, accompanied by design ideas for crafting them, and feel free to enrich them with your wisdom and experience:\\
        \hspace{6mm}- You can design questions to list the sequence between any selected two or more key actions, but DO NOT copy the quotation marks in the original text information when refer the actions.\\
        \hspace{6mm}- You can just randomly select one of the key action(s) and ask what are the action(s) before and after it to check whether the students have watched and understood the whole video. To answer the question student must refer to the video and compare different parts of the video.\\
        \hspace{6mm}- You can ask students summarize the key actions between randomly slected two actions.\\
    \hspace{3mm}- Refer naturally to key actions info using words such as \texttt{"before"}, \texttt{"earlier"}, \texttt{"prior to"}, \texttt{"following"}, \texttt{"after"}, \texttt{"subsequently"}, \texttt{"finally"}, etc., Do Not Mention \texttt{"action"}, \texttt{"step"} or \texttt{"steps"} when asking questions.\\
    \hspace{3mm}- Since the number of individuals in the video is unknown, avoid using specific pronouns like \texttt{"he"}, \texttt{"she"}, \texttt{"we"}, \texttt{"they"}, \texttt{"you"} etc. when asking the question. \\
    \hspace{3mm}- Note: the \texttt{"video\_info"} may not cover all the information in the video. Therefore, DO NOT USE absolute ordinal numbers like \texttt{"first thing"}, \texttt{"first"}, \texttt{"last"}, \texttt{"second"}, \texttt{"third"} etc. or other absolute words like \texttt{"only"}, \texttt{"beginning"}, \texttt{"final"}, \texttt{"end"}, etc. when asking questions or answering. \\
    \hspace{3mm}- DO NOT MENTION \texttt{"video\_info"}, \texttt{"video info"}, \texttt{"step"}, \texttt{"steps"}, \texttt{"key step"} etc. Ask and Answer it vividly and naturally as if you were watching the video. \\
    \hspace{3mm}- DO NOT include ``time interval'' ``timestamp'' or any number realted to the specific time in your answers or questions. Ask and Answer it natrually as if you were watching the video.\\
    \hspace{3mm}- ONLY ASK specific question that have definite answers. DO NOT ASK any question that cannot be answered confidently with the information from the provided video text info. DO NOT ASK ambiguous questions, DO NOT ASK overly generalized questions that may leave users uncertain about how to respond. \\
    \hspace{3mm}- Give a consice, brief reference Answer to each question, each Answer should be under 20 words. And your answer must be a direct reference to the information provided. DO NOT include anything that wasn't in the original video info.\\
    \hspace{3mm}- Return a list which include three question-answer pairs and your designed reason include explaining the question type and where the answer comes from, and each pair is in json format: \{\texttt{"Question"}: xxx, \texttt{"Answer"}: xxx, \texttt{"Reason"}: xxx\}.\\
    \end{tabular}
\end{tcolorbox}
\caption{Prompt template designed for generating \textbf{Coarse-Grained} temporal reasoning questions for videos in the \textbf{HiREST} dataset. The content highlighted in \textcolor{DeepBlue}{blue} changes according to the specific video.}
    \label{tab:Prompt Template for Hirest Coarse-Grained Q&A Generation}
\end{minipage}
\end{table*}

\begin{table*}[h!]
\centering
\begin{minipage}{0.95\textwidth}
\centering
\begin{tcolorbox}
    \centering
    \small
    \begin{tabular}{p{0.95\textwidth}}
    \textcolor{blue}{\textbf{System message}} \\
    You are an experienced teacher who are good at understanding long instructional video and capturing detailed objects appears in the video and extracting their relevant detailed info.
    Your task is to test students' ability to capture the detailed objects and extract their related objects and actions. \\
    \midrule
    \textcolor{blue}{\textbf{Prompt}} \\
    \#\# Instruction:\\
        \hspace{3mm}- User will give you video information in text, including the theme of the video(\texttt{"video theme"}), video's duration(\texttt{"duration"}), and key action within specific time intervals.\\
        \hspace{3mm}- Watch the video through the text video info carefully, and design three Fine-Grained, detailed, challenging question-answer pairs as if you are watching the video.\\
    \#\#  Input:\\
        \hspace{3mm}\textcolor{DeepBlue}{- video\_info:}\\
        \hspace{3mm}\textcolor{DeepBlue}{- video theme: Make a Plaster Mask}\\
        \hspace{3mm}\textcolor{DeepBlue}{- duration: 103.0s}\\
        \hspace{3mm}\textcolor{DeepBlue}{- key action within specific time intervals:}\\
            \hspace{6mm}\textcolor{DeepBlue}{- 35s $\sim$ 48s : clean out the face}\\
            \hspace{6mm}\textcolor{DeepBlue}{- 48s $\sim$ 60s : apply tissue using water on face}\\
            \hspace{6mm}\textcolor{DeepBlue}{- 60s $\sim$ 72s : apply it for full face}\\
            \hspace{6mm}\textcolor{DeepBlue}{- 72s $\sim$ 77s : put it under the neck}\\
            \hspace{6mm}\textcolor{DeepBlue}{- 77s $\sim$ 79s : dry it out}\\
    \#\#  Output:\\
        \hspace{3mm}- Use your extensive experience as much as possible to design three diverse, Fine-Grained, detailed, challenging questions that fully test students' ability to capture the detailed objects and extract the related objects' action after watching this long video. Below are some question types you can refer, accompanied by design ideas for crafting them:\\
            \hspace{6mm}- DO NOT ASK coarsed questions based on action, but ask detailed questions based on the objects.\\
            \hspace{6mm}- You can ask what object(s) are included in an action(s). To answer the question, student should find the action, and extract the object info.\\
            \hspace{6mm}- You can ask about the detailed attribute info with the objects you found, such as its appearance, color, size, its position, its type, etc. To answer the question, students need to watch the video carefully, and find the object, and recognize its detailed attributes.\\
            \hspace{6mm}- You can ask about the interactions associated with the objects you found, including the related objects and the related actions. To answer the question, student need carefully watch the video, find the object, extract the interaction info of the object. But DO NOT USE the words \texttt{"objects"} and \texttt{"interactions"} in your questions, ask questions naturally and vividly as if you were watching a video. \\
        \hspace{3mm}- Since the number of individuals in the video is unknown, avoid using specific pronouns like \texttt{"he"}, \texttt{"she"}, \texttt{"we"}, \texttt{"they"}, \texttt{"you"} etc. when asking the question. \\
        \hspace{3mm}- Note: the \texttt{"video\_info"} may not cover all the information in the video. Therefore, DO NOT USE absolute ordinal numbers like \texttt{"first thing"}, \texttt{"first"}, \texttt{"last"}, \texttt{"second"},\texttt{"third"} etc. or other absolute words like \texttt{"only"}, \texttt{"beginning"}, \texttt{"final"}, \texttt{"end"} etc. when asking questions or answering. \\
        \hspace{3mm}- DO NOT MENTION \texttt{"video\_info"}, \texttt{"video info"}, \texttt{"step"}, \texttt{"steps"}, \texttt{"key step"} etc. Ask and Answer it vividly and naturally as if you were watching the video. \\
        \hspace{3mm}- DO NOT include \texttt{"time interval"}, \texttt{"timestamp"} or any number related to the specific time in your answers or questions. Ask and Answer it naturally as if you were watching the video.\\
        \hspace{3mm}- ONLY ASK specific questions that have definite answers. DO NOT ASK any question that cannot be answered confidently with the information from the provided video text info. DO NOT ASK ambiguous questions, DO NOT ASK overly generalized questions that may leave users uncertain about how to respond. \\
        \hspace{3mm}- Give a concise, brief reference Answer to each question, each Answer should be under 20 words. And your answer must be a direct reference to the information provided. DO NOT include anything that wasn't in the original video info.\\
        \hspace{3mm}- Return a list which includes three question-answer pairs and your designed reason include explaining the question type and where the answer comes from, and each pair is in json format: \{\texttt{"Question"}: xxx, \texttt{"Answer"}: xxx, \texttt{"Reason"}: xxx\}.\\
    \end{tabular}
\end{tcolorbox}
\caption{Prompt template designed for generating \textbf{Fine-Grained} temporal reasoning questions for videos in the \textbf{HiREST} dataset. The content highlighted in \textcolor{DeepBlue}{blue} changes according to the specific video.}
    \label{tab:Prompt Template for Hirest Fine-Grained Q&A Generation}
\end{minipage}
\end{table*}

\begin{table*}[h!]
\centering
\begin{minipage}{0.95\textwidth}
\centering
\begin{tcolorbox}
    \centering
    \small
    \begin{tabular}{p{0.95\textwidth}}
    \textcolor{blue}{\textbf{System message}} \\
    You are an experienced teacher who are good at understanding long instructional video, capturing the action and movement of the video, and deeply understanding the meaning behind these movements.
    Your task is to test students' ability to capture action sequence and Fine-Grained action after watching this long video. \\
    \midrule
    \textcolor{blue}{\textbf{Prompt}} \\
    \#\# Instruction:\\
        \hspace{3mm}- User will give you video information in text, including the recipe type of the video(\texttt{"recipe type"}), video's duration(\texttt{"duration"}), and key action within specific time intervals.\\
        \hspace{3mm}- Watch the video through the text video info carefully, and design three inferential, challenging question-answer pairs as if you are watching the video.\\
    \#\#  Input:\\
        \hspace{3mm}\textcolor{DeepBlue}{- video\_info:}\\
        \hspace{3mm}\textcolor{DeepBlue}{- recipe type: grilled cheese}\\
        \hspace{3mm}\textcolor{DeepBlue}{- duration: 241.62s}\\
        \hspace{3mm}\textcolor{DeepBlue}{- key action within specific time intervals:}\\
            \hspace{6mm}\textcolor{DeepBlue}{- 90s $\sim$ 102s : spread margarine on two slices of white bread}\\
            \hspace{6mm}\textcolor{DeepBlue}{- 114s $\sim$ 127s : place a slice of cheese on the bread}\\
            \hspace{6mm}\textcolor{DeepBlue}{- 132s $\sim$ 138s : place the bread slices on top of each other and place in a hot pan}\\
            \hspace{6mm}\textcolor{DeepBlue}{- 139s $\sim$ 145s : flip the sandwich over and press down}\\
            \hspace{6mm}\textcolor{DeepBlue}{- 173s $\sim$ 174s : cut the sandwich in half diagonally}\\
    \#\#  Output:\\
        \hspace{3mm}- Use your extensive experience as much as possible to design three diverse, challenging, inferential questions that fully test students' ability to capture action sequence and Fine-Grained action after watching this long video. Below are some question types you can refer, accompanied by design ideas for crafting them, and feel free to enrich them with your wisdom and experience:\\
            \hspace{6mm}- You can design questions to list the sequence between any selected two or more key actions, but DO NOT copy the quotation marks in the original text information when refer the actions.\\
            \hspace{6mm}- You can just randomly select one of the key action(s) and ask what are the action(s) before and after it to check whether the students have watched and understood the whole video. To answer the question student must refer to the video and compare different parts of the video.\\
            \hspace{6mm}- You can ask students summarize the key actions between randomly slected two actions.\\
        \hspace{3mm}- Refer naturally to key actions info using words such as \texttt{"before"}, \texttt{"earlier"}, \texttt{"prior to"}, \texttt{"following"}, \texttt{"after"}, \texttt{"subsequently"}, \texttt{"finally"}, etc., Do Not Mention \texttt{"action"} \texttt{"step"} or \texttt{"steps"}  when asking questions.\\
        \hspace{3mm}- Since the number of individuals in the video is unknown, avoid using specific pronouns like "he", "she", "we", "they" "you" etc. when asking the question.\\
        \hspace{3mm}- Note: the \texttt{"video\_info"} may not cover all the information in the video. Therefore, DO NOT USE absolute ordinal numbers like \texttt{"first thing"}, \texttt{"first"}, \texttt{"last"}, \texttt{"second"},\texttt{"third"} etc. or other absolute words like \texttt{"only"},\texttt{"beginning"}, \texttt{"final"}, \texttt{"end"} etc. when asking questions or answering. \\
        \hspace{3mm}- DO NOT MENTION \texttt{"video\_info"}, \texttt{"video info"}, \texttt{"step"}, \texttt{"steps"}, \texttt{"key step"} etc. Ask and Answer it vividly and naturally as if you were watching the video. \\
        \hspace{3mm}- DO NOT include \texttt{"time interval"}, \texttt{"timestamp"} or any number related to the specific time in your answers or questions. Ask and Answer it naturally as if you were watching the video.\\
        \hspace{3mm}- ONLY ASK specific questions that have definite answers. DO NOT ASK any question that cannot be answered confidently with the information from the provided video text info. DO NOT ASK ambiguous questions, DO NOT ASK overly generalized questions that may leave users uncertain about how to respond. \\
        \hspace{3mm}- Give a concise, brief reference Answer to each question, each Answer should be under 20 words. And your answer must be a direct reference to the information provided. DO NOT include anything that wasn't in the original video info.\\
        \hspace{3mm}- Return a list which includes three question-answer pairs and your designed reason include explaining the question type and where the answer comes from, and each pair is in json format: \{\texttt{"Question"}: xxx, \texttt{"Answer"}: xxx, \texttt{"Reason"}: xxx\}.\\
    \end{tabular}
\end{tcolorbox}
\caption{Prompt template designed for generating \textbf{Coarse-Grained} temporal reasoning questions for videos in the \textbf{YouCook2} dataset. The content highlighted in \textcolor{DeepBlue}{blue} changes according to the specific video.}
    \label{tab:Prompt Template for YouCook2 Coarse-Grained Q&A Generation}
\end{minipage}
\end{table*}

\begin{table*}[h!]
\centering
\begin{minipage}{0.95\textwidth}
\centering
\begin{tcolorbox}
    \centering
    \small
    \begin{tabular}{p{0.95\textwidth}}
    \textcolor{blue}{\textbf{System message}} \\
    You are an experienced teacher who are good at understanding long instructional video and capturing detailed objects appears in the video and extracting their relevant detailed info.
    Your task is to test students' ability to capture the detailed objects and extract their related objects and actions. \\
    \midrule
    \textcolor{blue}{\textbf{Prompt}} \\
    \#\# Instruction:\\
        \hspace{3mm}- User will give you video information in text, including the recipe type of the video(\texttt{"recipe type"}), video's duration(\texttt{"duration"}), and key action within specific time intervals.\\
        \hspace{3mm}- Watch the video through the text video info carefully, and design three Fine-Grained, detailed, challenging question-answer pairs as if you are watching the video.\\
    \#\#  Input:\\
        \hspace{3mm}\textcolor{DeepBlue}{- video\_info:}\\
        \hspace{3mm}\textcolor{DeepBlue}{- recipe type: grilled cheese}\\
        \hspace{3mm}\textcolor{DeepBlue}{- duration: 241.62s}\\
        \hspace{3mm}\textcolor{DeepBlue}{- key action within specific time intervals:}\\
            \hspace{6mm}\textcolor{DeepBlue}{- 90s $\sim$ 102s : spread margarine on two slices of white bread}\\
            \hspace{6mm}\textcolor{DeepBlue}{- 114s $\sim$ 127s : place a slice of cheese on the bread}\\
            \hspace{6mm}\textcolor{DeepBlue}{- 132s $\sim$ 138s : place the bread slices on top of each other and place in a hot pan}\\
            \hspace{6mm}\textcolor{DeepBlue}{- 139s $\sim$ 145s : flip the sandwich over and press down}\\
            \hspace{6mm}\textcolor{DeepBlue}{- 173s $\sim$ 174s : cut the sandwich in half diagonally}\\
    \#\#  Output:\\
        \hspace{3mm}- Use your extensive experience as much as possible to design three diverse, Fine-Grained, detailed, challenging questions that fully test students' ability to capture the detailed objects and extract the related objects' action after watching this long video. Below are some question types you can refer, accompanied by design ideas for crafting them:\\
            \hspace{6mm}- DO NOT ASK coarsed questions based on action, but ask detailed questions based on the objects.\\
            \hspace{6mm}- You can ask what object(s) are included in an action(s). To answer the question, student should find the action, and extract the object info.\\
            \hspace{6mm}- You can ask about the detailed attribute info with the objects you found, such as its appearance, color, size, its position, its type, etc. To answer the question, students need to watch the video carefully, and find the object, and recognize its detailed attributes.\\
            \hspace{6mm}- You can ask about the interactions associated with the objects you found, including the related objects and the related actions. To answer the question, student need carefully watch the video, find the object, extract the interaction info of the object. But DO NOT USE the words \texttt{"objects"} and \texttt{"interactions"} in your questions, ask questions naturally and vividly as if you were watching a video. \\
        \hspace{3mm}- Since the number of individuals in the video is unknown, avoid using specific pronouns like \texttt{"he"}, \texttt{"she"}, \texttt{"we"}, \texttt{"they"} \texttt{"you"} etc. when asking the question. \\
        \hspace{3mm}- Note: the \texttt{"video\_info"} may not cover all the information in the video. Therefore, DO NOT USE absolute ordinal numbers like \texttt{"first thing"}, \texttt{"first"}, \texttt{"last"}, \texttt{"second"},\texttt{"third"} etc. or other absolute words like \texttt{"only"},\texttt{"beginning"}, \texttt{"final"}, \texttt{"end"} etc. when asking questions or answering. \\
        \hspace{3mm}- DO NOT MENTION \texttt{"video\_info"}, \texttt{"video info"}, \texttt{"step"}, \texttt{"steps"}, \texttt{"key step"} etc. Ask and Answer it vividly and naturally as if you were watching the video. \\
        \hspace{3mm}- DO NOT include \texttt{"time interval"}, \texttt{"timestamp"} or any number related to the specific time in your answers or questions. Ask and Answer it naturally as if you were watching the video.\\
        \hspace{3mm}- ONLY ASK specific questions that have definite answers. DO NOT ASK any question that cannot be answered confidently with the information from the provided video text info. DO NOT ASK ambiguous questions, DO NOT ASK overly generalized questions that may leave users uncertain about how to respond. \\
        \hspace{3mm}- Give a concise, brief reference Answer to each question, each Answer should be under 20 words. And your answer must be a direct reference to the information provided. DO NOT include anything that wasn't in the original video info.\\
        \hspace{3mm}- Return a list which includes three question-answer pairs and your designed reason include explaining the question type and where the answer comes from, and each pair is in json format: \{\texttt{"Question"}: xxx, \texttt{"Answer"}: xxx, \texttt{"Reason"}: xxx\}.\\
    \end{tabular}
\end{tcolorbox}
\caption{Prompt template designed for generating \textbf{Fine-Grained} temporal reasoning questions for videos in the \textbf{YouCook2} dataset. The content highlighted in \textcolor{DeepBlue}{blue} changes according to the specific video.}
\label{tab:Prompt Template for YouCook2 Fine-Grained Q&A Generation}
\end{minipage}
\end{table*}

\begin{table*}[h!]
\centering
\begin{minipage}{0.95\textwidth}
\centering
\begin{tcolorbox}
    \centering
    \small
    \begin{tabular}{p{0.95\textwidth}}
    \textcolor{blue}{\textbf{System message}} \\
    You excel in interpreting video footage recorded from a first-person perspective by a pair of smart glasses, where an Augmented Reality (AR) assistant is integrated. Your skill set includes capturing the actions sequence and Fine-Grained actions of the video. Your task is to anticipate potential user questions and provide relevant and helpful responses. In this task, you will play two roles: \\
    \hspace{3mm}- The User: Imagine you're wearing smart glasses, and you want to recall past actions, actions sequence or predict what will happen next. Ask questions that recall past actions, actions sequence or predict future actions in a vivid and natural manner, as if you were really there.\\
    \hspace{3mm}- The Augmented Reality (AR) assistant: Answer the user's questions. Your answers should be useful and should be based on your understanding of the main actions and action sequences in the video. \\
    \midrule
    \textcolor{blue}{\textbf{Prompt}} \\
    \#\# Instruction:\\
        \hspace{3mm}- I'll provide you with text-based video clip info, which is a segment extracted from an instructional video.\\
        \hspace{3mm}- The video clip info includes the video's goal(\texttt{"video goal"}), duration(\texttt{"duration"}), and \texttt{"key action within specific time intervals"}, which are divided into steps and substeps, shown by different indents.\\
        \hspace{3mm}- Thoroughly examine the text video info and design five inferential, challenging question-answer pairs. As if you were watching the video, ask an AR assistant. Then, as the AR assistant, provide useful response.\\
    \#\#  Input:\\
        \hspace{3mm}\textcolor{DeepBlue}{- Instructional Video Clip Info(a segment extracted from the original video):}\\
        \hspace{6mm}\textcolor{DeepBlue}{- video goal: making pan seared meal}\\
        \hspace{6mm}\textcolor{DeepBlue}{- key action within specific time intervals(are divided into steps and substeps, shown by different indents):}\\
            \hspace{9mm}\textcolor{DeepBlue}{- 0.0s $\sim$ 27.16s: Add butter to a pan(heat butter)}\\
                \hspace{12mm}\textcolor{DeepBlue}{- (0.16s $\sim$ 18.61s: Add oil to a pan(add butter to skillet))}\\
                \hspace{12mm}\textcolor{DeepBlue}{- (18.61s $\sim$ 27.2s: Store ingredients in refrigerator or freezer(put butter in refrigerator))}\\
            \hspace{9mm}\textcolor{DeepBlue}{- ...... }\\
            \hspace{9mm}\textcolor{DeepBlue}{- 273.64s $\sim$ 319.18s: Make egg salad(make egg salad)}\\
                \hspace{12mm}\textcolor{DeepBlue}{- (273.72s $\sim$ 299.58s: Cut onion(cut onion on kitchen board))}\\
                \hspace{12mm}\textcolor{DeepBlue}{- (299.58s $\sim$ 310.41s: Add onion to recipe(add chopped onions to plate))}\\
    \#\#  Output:\\
        \hspace{3mm}- Use your extensive experience as much as possible to design five diverse, challenging, inferential questions that fully test the AR assistant's ability to capture actions sequence and Fine-Grained actions after watching this video clip. Below are some question types you can refer, accompanied by design ideas for crafting them, and feel free to enrich them with your wisdom and experience:\\
            \hspace{6mm}- You can design questions to list the sequence between any selected two or more key actions, but DO NOT directly copy the punctuation marks from the given info when referring to the actions.\\
            \hspace{6mm}- You can just randomly select one of the key action(s) and ask what are the action(s) before and after it to check whether the assistants have watched and understood the whole video. To answer question the assistant must refer to the video and compare different parts of the video.\\
            \hspace{6mm}- You can ask the assistant summarize the key actions between randomly slected two actions.\\
        \hspace{3mm}- Refer naturally to key actions info using words such as \texttt{"before"}, \texttt{"earlier"}, \texttt{"prior to"}, \texttt{"following"}, \texttt{"after"}, \texttt{"subsequently"}, \texttt{"finally"}, etc., Do Not Mention \texttt{"action"} \texttt{"step"} or \texttt{"steps"} when asking questions.\\
        \hspace{3mm}- Note: the \texttt{"Instructional Video Clip Info"} do not cover all the information in the video. Therefore, DO NOT USE absolute ordinal numbers like \texttt{"first thing"}, \texttt{"first"}, \texttt{"last"}, \texttt{"second"},\texttt{"third"} etc. or other absolute words like \texttt{"only"},\texttt{"beginning"}, \texttt{"final"}, \texttt{"end"} etc. when asking questions or answering. \\
        \hspace{3mm}- DO NOT MENTION \texttt{"video info"}, \texttt{"step"}, \texttt{"substep"}, \texttt{"subaction"}, \texttt{"the individual"}, \texttt{"the person"} etc. Pose questions and provide answers in a vivid and natural way, as if you were a user interacting with the AR assistant while doing daily activities, and as if you were the AR assistant responding to the user's inquiries. \\
        \hspace{3mm}- DO NOT include \texttt{"time interval"}, \texttt{"timestamp"} or any number related to the specific time in your answers or questions.\\
        \hspace{3mm}- ONLY ASK specific questions that have definite answers. DO NOT ASK any question that cannot be answered confidently with the information from the provided video text info. DO NOT ASK ambiguous questions, DO NOT ASK overly generalized questions that may leave the assistant uncertain about how to respond.\\
        \hspace{3mm}- Focus on the specificity and challenge of each question. Avoid questions that are too broad or too simple.\\
        \hspace{3mm}- Please ensure that the questions you generate are not repetitive and cover as much of the video content as possible.\\
        \hspace{3mm}- Give a concise, brief reference answer to each question, each Answer should be under 20 words. And your answer must be a direct reference to the information provided. DO NOT include anything that wasn't in the original video info.\\
        \hspace{3mm}- Return a list that includes five question-answer pairs, and for each pair, provide a designed reason that explains the question type and where the answer comes from. Each pair should be in JSON format: \{\texttt{"Question"}: xxx, \texttt{"Answer"}: xxx, \texttt{"Reason"}: xxx\}.\\
    \end{tabular}
\end{tcolorbox}
\caption{Prompt template designed for generating \textbf{Coarse-Grained} temporal reasoning questions for videos in the \textbf{Ego4D Goal-Step} dataset. The content highlighted in \textcolor{DeepBlue}{blue} changes according to the specific video.}
    \label{tab:Prompt Template for Ego4D Goal-Step Coarse-Grained Q&A Generation}
\end{minipage}
\end{table*}
\begin{table*}[h!]
\centering
\begin{minipage}{0.95\textwidth}
\centering
\begin{tcolorbox}
    \centering
    \small
    \begin{tabular}{p{0.95\textwidth}}
    \textcolor{blue}{\textbf{System message}} \\
    Define the role and expertise of the AI model. Clearly state the task, the context, and the specific skills required. \\
    Example: \\
    \textit{You excel in interpreting video footage recorded from a first-person perspective by a pair of smart glasses, where an Augmented Reality (AR) assistant is integrated. Your skill set includes capturing the action sequences and fine-grained actions of the video. Your task is to anticipate potential user questions and provide relevant and helpful responses.} \\
    \midrule
    \textcolor{blue}{\textbf{Roles}} \\
    Specify the roles the AI model will play (e.g., user and AR assistant). \\
    Example: \\
    \textit{- The User: Imagine you're wearing smart glasses, and you want to recall past actions, action sequences, or predict what will happen next. Ask questions that recall past actions, action sequences, or predict future actions in a vivid and natural manner, as if you were really there.} \\
    \textit{- The Augmented Reality (AR) Assistant: Answer the user's questions. Your answers should be useful and based on your understanding of the main actions and action sequences in the video.} \\
    \midrule
    \textcolor{blue}{\textbf{Prompt Structure}} \\
    Organize the prompt into clear sections: \textbf{Instruction}, \textbf{Input}, and \textbf{Output}. \\
    \textbf{Instruction}: \\
    \textit{- Thoroughly examine the provided video clip info and design inferential, challenging question-answer pairs.} \\
    \textit{- As the user, ask questions as if you were interacting with the AR assistant.} \\
    \textit{- As the AR assistant, provide useful responses based on the video content.} \\
    \textbf{Input}: \\
    \textit{- Instructional Video Clip Info: A segment extracted from the original video, including:} \\
    \hspace{3mm}\textit{- Video Goal: The main objective of the video.} \\
    \hspace{3mm}\textit{- Key Actions Within Specific Time Intervals: Divided into steps and substeps, shown by different indents.} \\
    \textbf{Output}: \\
    \textit{- Design five diverse, challenging, inferential questions that test the AR assistant's ability to capture action sequences and fine-grained actions.} \\
    \textit{- Provide concise, brief answers to each question (under 20 words).} \\
    \textit{- Return a list of question-answer pairs in JSON format:} \\
    \hspace{3mm}\textit{\{"Question": xxx, "Answer": xxx, "Reason": xxx\}.} \\
    \midrule
    \textcolor{blue}{\textbf{Constraints and Guidelines}} \\
    Include specific rules to ensure the output is consistent and aligned with the task. \\
    Example: \\
    \textit{- Do Not Use Absolute Terms: Avoid words like "first," "last," "only," "beginning," or "end."} \\
    \textit{- Do Not Mention Technical Terms: Avoid words like "action," "step," "substep," or "video info."} \\
    \textit{- Focus on Specificity: Ask questions that have definite answers based on the provided video content.} \\
    \textit{- Avoid Ambiguity: Do not ask overly generalized or ambiguous questions.} \\
    \textit{- Refer Naturally: Use words like "before," "after," "earlier," "following," etc., to refer to actions.} \\
    \end{tabular}
\end{tcolorbox}
\caption{Generalized Prompt Template for Video Analysis Tasks. The content highlighted in \textcolor{DeepBlue}{blue} changes according to the specific task.}
\label{tab:Generalized Prompt Template for Video Analysis Tasks}
\end{minipage}
\end{table*}

\begin{table*}[h!]
\centering
\begin{minipage}{0.95\textwidth}
\centering
\begin{tcolorbox}
    \centering
    \small
    \begin{tabular}{p{0.95\textwidth}}
    \textcolor{blue}{\textbf{System message}} \\
    \textbf{Task Granularity}: \\
    - \textbf{Coarse-Grained}: You are an experienced teacher who is good at understanding long instructional videos, capturing the action and movement of the video, and deeply understanding the meaning behind these movements. Your task is to test students’ ability to capture action sequences and fine-grained actions after watching this long video. \\
    - \textbf{Fine-Grained}: You are an experienced teacher who is good at understanding long instructional videos and capturing detailed objects that appear in the video, as well as extracting their relevant detailed information. Your task is to test students’ ability to capture detailed objects and extract their related objects and actions. \\
    \textbf{Dataset Perspective}: \\
    - \textbf{Third-Person}: You are an experienced teacher who is good at understanding long instructional videos and capturing detailed objects that appear in the video, as well as extracting their relevant detailed information. Your task is to test students’ ability to capture detailed objects and extract their related objects and actions. \\
    - \textbf{First-Person}: You excel in interpreting video footage recorded from a first-person perspective by a pair of smart glasses, where an Augmented Reality (AR) assistant is integrated. Your skill set includes capturing the action sequences and fine-grained actions of the video. Your task is to anticipate potential user questions and provide relevant and helpful responses. In this task, you will play two roles: \\
    \hspace{3mm}- \textbf{The User}: Imagine you’re wearing smart glasses, and you want to recall past actions, action sequences, or predict what will happen next. Ask questions that recall past actions, action sequences, or predict future actions in a vivid and natural manner, as if you were really there. \\
    \hspace{3mm}- \textbf{The Augmented Reality (AR) Assistant}: Answer the user’s questions. Your answers should be useful and should be based on your understanding of the main actions and action sequences in the video. \\
    \midrule
    \textcolor{blue}{\textbf{Prompt Structure}} \\
    \textbf{Instruction}: \\
    - Thoroughly examine the provided video clip info and design inferential, challenging question-answer pairs. \\
    - As the user, ask questions as if you were interacting with the AR assistant. \\
    - As the AR assistant, provide useful responses based on the video content. \\
    \textbf{Input}: \\
    - \textbf{Instructional Video Clip Info}: A segment extracted from the original video, including: \\
    \hspace{3mm}- \textbf{Video Goal}: The main objective of the video. \\
    \hspace{3mm}- \textbf{Key Actions Within Specific Time Intervals}: Divided into steps and substeps, shown by different indents. \\
    \textbf{Output}: \\
    - Design five diverse, challenging, inferential questions that test the AR assistant's ability to capture action sequences and fine-grained actions. \\
    - Provide concise, brief answers to each question (under 20 words). \\
    - Return a list of question-answer pairs in JSON format: \\
    \hspace{3mm}\{\texttt{"Question": xxx, "Answer": xxx, "Reason": xxx}. \\
    \midrule
    \textcolor{blue}{\textbf{Constraints and Guidelines}} \\
    - \textbf{Do Not Use Absolute Terms}: Avoid words like "first," "last," "only," "beginning," or "end." \\
    - \textbf{Do Not Mention Technical Terms}: Avoid words like "action," "step," "substep," or "video info." \\
    - \textbf{Focus on Specificity}: Ask questions that have definite answers based on the provided video content. \\
    - \textbf{Avoid Ambiguity}: Do not ask overly generalized or ambiguous questions. \\
    - \textbf{Refer Naturally}: Use words like "before," "after," "earlier," "following," etc., to refer to actions. \\
    \end{tabular}
\end{tcolorbox}
\caption{Generalized Prompt Template for Video Analysis Tasks. The content highlighted in \textcolor{DeepBlue}{blue} changes according to the specific task.}
\label{tab:Generalized Prompt Template for Video Analysis Tasks}
\end{minipage}
\end{table*}

\begin{table*}[h!]
\centering
\begin{minipage}{0.95\textwidth}
\centering
\begin{tcolorbox}
    \centering
    \small
    \begin{tabular}{p{0.95\textwidth}}
    \textcolor{blue}{\textbf{System message}} \\
    The System Message varies based on \textbf{task granularity} and \textbf{dataset perspective}. \\
    \textbf{Task Granularity}: \\
    - \textbf{Coarse-Grained}: \textit{You are an experienced teacher who is good at understanding long instructional videos, capturing the action and movement of the video, and deeply understanding the meaning behind these movements. Your task is to test students’ ability to capture action sequences and fine-grained actions after watching this long video.} \\
    - \textbf{Fine-Grained}: \textit{You are an experienced teacher who is good at understanding long instructional videos and capturing detailed objects that appear in the video, as well as extracting their relevant detailed information. Your task is to test students’ ability to capture detailed objects and extract their related objects and actions.} \\
    \textbf{Dataset Perspective}: \\
    - \textbf{Third-Person}: \textit{You are an experienced teacher who is good at understanding long instructional videos and capturing detailed objects that appear in the video, as well as extracting their relevant detailed information. Your task is to test students’ ability to capture detailed objects and extract their related objects and actions.} \\
    - \textbf{First-Person}: \textit{You excel in interpreting video footage recorded from a first-person perspective by a pair of smart glasses, where an Augmented Reality (AR) assistant is integrated. Your skill set includes capturing the action sequences and fine-grained actions of the video. Your task is to anticipate potential user questions and provide relevant and helpful responses. In this task, you will play two roles:} \\
    \hspace{3mm}- \textbf{The User}: \textit{Imagine you’re wearing smart glasses, and you want to recall past actions, action sequences, or predict what will happen next. Ask questions that recall past actions, action sequences, or predict future actions in a vivid and natural manner, as if you were really there.} \\
    \hspace{3mm}- \textbf{The Augmented Reality (AR) Assistant}: \textit{Answer the user’s questions. Your answers should be useful and should be based on your understanding of the main actions and action sequences in the video.} \\
    \midrule
    \textcolor{blue}{\textbf{Prompt Structure}} \\
    \textbf{Instruction}: \\
    - Thoroughly examine the provided video clip info and design inferential, challenging question-answer pairs. \\
    - As the user, ask questions as if you were interacting with the AR assistant. \\
    - As the AR assistant, provide useful responses based on the video content. \\
    \textbf{Input}: \\
    - \textbf{Instructional Video Clip Info}: A segment extracted from the original video, including: \\
    \hspace{3mm}- \textbf{Video Goal}: The main objective of the video. \\
    \hspace{3mm}- \textbf{Key Actions Within Specific Time Intervals}: Divided into steps and substeps, shown by different indents. \\
    \textbf{Output}: \\
    - Design five diverse, challenging, inferential questions that test the AR assistant's ability to capture action sequences and fine-grained actions. \\
    - Provide concise, brief answers to each question (under 20 words). \\
    - Return a list of question-answer pairs in JSON format: \\
    \hspace{3mm}\texttt{"Question": xxx, "Answer": xxx, "Reason": xxx}. \\
    \midrule
    \textcolor{blue}{\textbf{Constraints and Guidelines}} \\
    - \textbf{Do Not Use Absolute Terms}: Avoid words like "first," "last," "only," "beginning," or "end." \\
    - \textbf{Do Not Mention Technical Terms}: Avoid words like "action," "step," "substep," or "video info." \\
    - \textbf{Focus on Specificity}: Ask questions that have definite answers based on the provided video content. \\
    - \textbf{Avoid Ambiguity}: Do not ask overly generalized or ambiguous questions. \\
    - \textbf{Refer Naturally}: Use words like "before," "after," "earlier," "following," etc., to refer to actions. \\
    \end{tabular}
\end{tcolorbox}
\caption{Generalized Prompt Template for Video Analysis Tasks. The content highlighted in \textcolor{DeepBlue}{blue} changes according to the specific task. Examples in \textit{italics} illustrate variations based on task granularity and dataset perspective.}
\label{tab:Generalized Prompt Template for Video Analysis Tasks}
\end{minipage}
\end{table*}

\begin{table*}[h!]
\centering
\begin{minipage}{0.95\textwidth}
\centering
\begin{tcolorbox}
    \centering
    \small
    \begin{tabular}{>{\raggedright\arraybackslash}p{0.95\textwidth}}
    \textcolor{blue}{\textbf{System message}} \\
    The System Message varies based on \textbf{task granularity}. \\
    \textbf{Task Granularity}: \\
    - \textbf{Coarse-Grained}: You are an experienced teacher who excels in understanding long instructional videos, capturing the action and movement, and deeply interpreting these actions. Your task is to test the ability to \textbf{capture action sequences and fine-grained actions} after viewing the instructional video. \\
    - \textbf{Fine-Grained}: You are an experienced teacher who excels in understanding long instructional videos and capturing detailed objects that appear in the video, as well as extracting their relevant detailed information. Your task is to test the ability to capture \textbf{detailed objects and extract their related actions and objects}. \\
    \midrule
    \textcolor{blue}{\textbf{Prompt Structure}} \\
    \textbf{Instruction}: \\
    - Thoroughly examine the provided video clip info and design challenging question-answer pairs. \\
    - As the user, ask questions as if interacting with the AR assistant. \\
    - As the AR assistant, provide useful responses based on the video content. \\
    \textbf{Input}: \\
    - \textbf{Instructional Video Clip Info}: A segment extracted from the original video, including: \\
    \hspace{3mm}- \textbf{Video Goal}: The main objective of the video. \\
    \hspace{3mm}- \textbf{Key Actions Within Specific Time Intervals}: Divided into steps and substeps, indicated by different indents. \\
    \textbf{Output}: \\
    - Design five diverse, challenging, inferential questions that test the AR assistant's ability to capture action sequences and fine-grained actions. \\
    - Provide concise, brief answers to each question (under 20 words). \\
    - Return a list of question-answer pairs in JSON format: \\
    \hspace{3mm}\texttt{"Question": xxx, "Answer": xxx, "Reason": xxx}. \\
    \midrule
    \textcolor{blue}{\textbf{Constraints and Guidelines}} \\
    - \textbf{Do Not Use Absolute Terms}: Avoid words like "first," "last," "only," "beginning," or "end." \\
    - \textbf{Do Not Mention Technical Terms}: Avoid words like "action," "step," "substep," or "video info." \\
    - \textbf{Focus on Specificity}: Ask questions that have definite answers based on the provided video content. \\
    - \textbf{Avoid Ambiguity}: Do not ask overly generalized or ambiguous questions. \\
    - \textbf{Refer Naturally}: Use words like "before," "after," "earlier," "following," etc., to refer to actions. \\
    \end{tabular}
\end{tcolorbox}
\caption{Generalized Prompt Template for Video Analysis Tasks. The content highlighted in \textcolor{blue}{blue} changes according to the specific task.}
\label{tab:Generalized Prompt Template for Video Analysis Tasks}
\end{minipage}
\end{table*}

\begin{table*}[h!]
\centering
\begin{minipage}{0.95\textwidth}
\centering
\begin{tcolorbox}
    \centering
    \small
    \begin{tabular}{>{\raggedright\arraybackslash}p{0.95\textwidth}}
    \textcolor{blue}{\textbf{System Message}} \\
    \textbf{Purpose}: Sets the context and role for the task, defining the system's behavior and task nature. \\
    \textit{Example}: The system message is divided into two levels based on task granularity: \\
    - \textbf{Coarse-Grained}: Focuses on capturing the overall action sequences in the video. \\
    - \textbf{Fine-Grained}: Emphasizes identifying detailed objects and extracting their related information. This distinction ensures the generated prompts align with task requirements. \\
    \midrule
    \textcolor{blue}{\textbf{Prompt Structure}} \\
    \textbf{Purpose}: Provides clear operational guidelines for generating prompts. \\
    \textit{Example}: GPT needs to analyze video content, design challenging QA pairs, and simulate user-AR assistant interactions. Specifically: \\
    - \textbf{Instruction}: Guides the task by requiring in-depth analysis of video content and the design of challenging questions and answers. \\
    - \textbf{Input}: Defines the input format and content, including the video goal and key actions (divided into steps and substeps). This ensures the model has sufficient context to understand the video and generate appropriate responses. \\
    - \textbf{Output}: Specifies the output format and requirements, such as generating five diverse QA pairs in JSON format. This ensures the output is structured, concise, and meets task needs. \\
    \midrule
    \textcolor{blue}{\textbf{Constraints and Guidelines}} \\
    \textbf{Purpose}: Provides rules and limitations to ensure clarity, specificity, and naturalness in the generated content. \\
    \textit{Example}: GPT must adhere to the following rules: \\
    - Avoid absolute terms (e.g., "first," "last") and technical terms (e.g., "action," "step"). \\
    - Focus on specificity and avoid ambiguity in questions. \\
    - Use natural temporal references (e.g., "before," "after") to refer to actions. \\
    \end{tabular}
\end{tcolorbox}
\caption{Generalized Prompt Template for Video Analysis Tasks. The content highlighted in \textcolor{blue}{blue} changes according to the specific task.}
\label{tab:Generalized Prompt Template for Video Analysis Tasks}
\end{minipage}
\end{table*}

\begin{table*}[h!]
\centering
\begin{minipage}{0.95\textwidth}
\centering
\begin{tcolorbox}
    \centering
    \small
    \begin{tabular}{>{\raggedright\arraybackslash}p{0.95\textwidth}}
    \textcolor{blue}{\textbf{System Message}} \\
    \textbf{Purpose}: Sets the context and role for the task, defining the system's behavior and task nature. \\
    \textit{Example}: The system message is divided into two levels based on task granularity: \\
    - \textbf{Coarse-Grained}: Focuses on capturing the overall action sequences in the video. \\
    - \textbf{Fine-Grained}: Emphasizes identifying detailed objects and extracting their related information. \\
    \midrule
    \textcolor{blue}{\textbf{Prompt Structure}} \\
    \textbf{Purpose}: Provides clear operational guidelines for generating prompts, ensuring the model understands the task requirements and generates appropriate outputs. \\
    \midrule
    \textcolor{blue}{\textbf{Constraints and Guidelines}} \\
    \textbf{Purpose}: Defines rules and limitations to ensure the generated content is clear, specific, and natural, avoiding ambiguity and technical jargon. \\
    \end{tabular}
\end{tcolorbox}
\caption{Generalized Prompt Template for Video Analysis Tasks. The content highlighted in \textcolor{blue}{blue} changes according to the specific task.}
\label{tab:Generalized Prompt Template for Video Analysis Tasks}
\end{minipage}
\end{table*}
\begin{table*}[h!]
\centering
\begin{minipage}{0.95\textwidth}
\centering
\begin{tcolorbox}
    \centering
    \small
    \begin{tabular}{>{\raggedright\arraybackslash}p{0.95\textwidth}}
    \textcolor{blue}{\textbf{System Message}} (Defines the context and role for the task, specifying the system's behavior and task nature.) \\
    Example: The system message is divided into two levels based on task granularity:
    - \textbf{Coarse-Grained} (Focuses on capturing the overall action sequences in the video, guiding GPT-4 to generate questions involving action sequence recognition and specific action localization.) \\
    - \textbf{Fine-Grained} (Emphasizes identifying detailed objects and extracting their related information, guiding GPT-4 to generate questions about the appearance of objects in the video while focusing on understanding dynamic actions.) \\
    \midrule
    \textcolor{blue}{\textbf{Prompt Structure}} (Provides clear operational guidelines for generating prompts, ensuring the model understands the task requirements and generates appropriate outputs.) \\
    - \#\# \textbf{Instruction} (Guides the task by requiring in-depth analysis of video content and the design of challenging questions and answers.) \\
    - \#\# \textbf{Input} (Defines the input format and content, including the video goal and key actions divided into steps and substeps.) \\
    - \#\# \textbf{Output} (Specifies the output format and requirements, such as generating five diverse QA pairs in JSON format.) \\
    \midrule
    \textcolor{blue}{\textbf{Constraints and Guidelines}} (Defines rules and limitations to ensure the generated content is clear, specific, and natural, avoiding ambiguity and technical jargon.) \\
    \end{tabular}
\end{tcolorbox}
\caption{Generalized Prompt Template for Video Analysis Tasks. The content highlighted in \textcolor{blue}{blue} changes according to the specific task.}
\label{tab:Generalized Prompt Template for Video Analysis Tasks}
\end{minipage}
\end{table*}

\begin{table*}[h!]
\centering
\begin{minipage}{0.95\textwidth}
\centering
\begin{tcolorbox}
    \centering
    \small
    \begin{tabular}{>{\raggedright\arraybackslash}p{0.95\textwidth}}
    \textcolor{blue}{\textbf{System Message}} (Defines the context and role for the task, specifying the system's behavior and task nature.) \\
    \textit{Example: The system message is divided into two levels based on task granularity:} \\
    \textit{- \textbf{Coarse-Grained} (Focuses on capturing the overall action sequences in the video, guiding GPT-4 to generate questions involving action sequence recognition and specific action localization.)} \\
    \textit{- \textbf{Fine-Grained} (Emphasizes identifying detailed objects and extracting their related information, guiding GPT-4 to generate questions about the appearance of objects in the video while focusing on understanding dynamic actions.)} \\
    \midrule
    \textcolor{blue}{\textbf{Prompt Structure}} (Provides clear operational guidelines for generating prompts, ensuring the model understands the task requirements and generates appropriate outputs.) \\
    - \textbf{\#\# Instruction} (Guides the task by requiring in-depth analysis of video content and the design of challenging questions and answers.) \\
    - \textbf{\#\# Input} (Defines the input format and content, including the video goal and key actions divided into steps and substeps.) \\
    - \textbf{\#\# Output} (Specifies the output format and requirements, such as generating five diverse QA pairs in JSON format.) \\
    \midrule
    \textcolor{blue}{\textbf{Constraints and Guidelines}} (Defines rules and limitations to ensure the generated content is clear, specific, and natural, avoiding ambiguity and technical jargon.) \\
    \end{tabular}
\end{tcolorbox}
\caption{Generalized Prompt Template for Video Analysis Tasks. The content highlighted in \textcolor{blue}{blue} changes according to the specific task.}
\label{tab:Generalized Prompt Template for Video Analysis Tasks}
\end{minipage}
\end{table*}

\begin{table*}[h!]
\centering
\begin{minipage}{0.95\textwidth}
\centering
\begin{tcolorbox}
    \centering
    \small
    \begin{tabular}{>{\raggedright\arraybackslash}p{0.95\textwidth}}
    \textcolor{blue}{\textbf{System Message}} (Defines the context and role for the task, specifying the system's behavior and task nature.) \\
    \textit{Example: The system message is divided into two levels based on task granularity:} \\
    \textit{- \textbf{Coarse-Grained} (Focuses on capturing the overall action sequences in the video, guiding GPT-4 to generate questions involving action sequence recognition and specific action localization.)} \\
    \textit{- \textbf{Fine-Grained} (Emphasizes identifying detailed objects and extracting their related information, guiding GPT-4 to generate questions about the appearance of objects in the video while focusing on understanding dynamic actions.)} \\
    \midrule
    \textcolor{blue}{\textbf{Prompt Structure}} (Provides clear operational guidelines for generating prompts, ensuring the model understands the task requirements and generates appropriate outputs.) \\
    - \textbf{Instruction} (Guides the task by requiring in-depth analysis of video content and the design of challenging questions and answers.) \\
    - \textbf{Input} (Defines the input format and content, including the video goal and key actions divided into steps and substeps.) \\
    \textit{Example:} \\
    \textit{\hspace{3mm}- Instructional Video Clip Info (a segment extracted from the original video):} \\
    \textit{\hspace{6mm}- Video Goal: Making pan seared meal} \\
    \textit{\hspace{6mm}- Key Actions Within Specific Time Intervals (divided into steps and substeps, shown by different indents):} \\
    \textit{\hspace{9mm}- 0.0s $\sim$ 27.16s: Add butter to a pan (heat butter)} \\
    \textit{\hspace{12mm}- (0.16s $\sim$ 18.61s: Add oil to a pan (add butter to skillet))} \\
    \textit{\hspace{12mm}- (18.61s $\sim$ 27.2s: Store ingredients in refrigerator or freezer (put butter in refrigerator))} \\
    \textit{\hspace{9mm}- ...... } \\
    \textit{\hspace{9mm}- 273.64s $\sim$ 319.18s: Make egg salad (make egg salad)} \\
    \textit{\hspace{12mm}- (273.72s $\sim$ 299.58s: Cut onion (cut onion on kitchen board))} \\
    \textit{\hspace{12mm}- (299.58s $\sim$ 310.41s: Add onion to recipe (add chopped onions to plate))} \\
    - \textbf{Output} (Specifies the output format and requirements, such as generating five diverse QA pairs in JSON format.) \\
    \midrule
    \textcolor{blue}{\textbf{Constraints and Guidelines}} (Defines rules and limitations to ensure the generated content is clear, specific, and natural, avoiding ambiguity and technical jargon.) \\
    \end{tabular}
\end{tcolorbox}
\caption{Generalized Prompt Template for Video Analysis Tasks. The content highlighted in \textcolor{blue}{blue} changes according to the specific task.}
\label{tab:Generalized Prompt Template for Video Analysis Tasks}
\end{minipage}
\end{table*}

\begin{table*}[h!]
\centering
\begin{minipage}{0.95\textwidth}
\centering
\begin{tcolorbox}
    \centering
    \small
    \begin{tabular}{>{\raggedright\arraybackslash}p{0.95\textwidth}}
    \textcolor{blue}{\textbf{System Message}} (Defines the context and role for the task, specifying the system's behavior and task nature.) \\
    \textit{Example: The system message is divided into two levels based on task granularity:} \\
    \textit{- \textbf{Coarse-Grained} (Focuses on capturing the overall action sequences in the video, guiding GPT-4 to generate questions involving action sequence recognition and specific action localization.)} \\
    \textit{- \textbf{Fine-Grained} (Emphasizes identifying detailed objects and extracting their related information, guiding GPT-4 to generate questions about the appearance of objects in the video while focusing on understanding dynamic actions.)} \\
    \midrule
    \textcolor{blue}{\textbf{Prompt}} (Provides clear operational guidelines for generating prompts, ensuring the model understands the task requirements and generates appropriate outputs.) \\
    \#\# Instruction (Guides the task by requiring in-depth analysis of video content and the design of challenging questions and answers.) \\
    \#\# Input (Defines the input format and content, including the video goal and key actions divided into steps and substeps.) \\
    \textit{Example:} \\
    \textit{\hspace{3mm}\textcolor{DeepBlue}{- Instructional Video Clip Info (a segment extracted from the original video):}} \\
    \textit{\hspace{6mm}\textcolor{DeepBlue}{- Video Goal: Making pan seared meal}} \\
    \textit{\hspace{6mm}\textcolor{DeepBlue}{- Key Actions Within Specific Time Intervals (divided into steps and substeps, shown by different indents):}} \\
    \textit{\hspace{9mm}\textcolor{DeepBlue}{- 0.0s $\sim$ 27.16s: Add butter to a pan (heat butter)}} \\
    \textit{\hspace{12mm}\textcolor{DeepBlue}{- (0.16s $\sim$ 18.61s: Add oil to a pan (add butter to skillet))}} \\
    \textit{\hspace{12mm}\textcolor{DeepBlue}{- (18.61s $\sim$ 27.2s: Store ingredients in refrigerator or freezer (put butter in refrigerator))}} \\
    \textit{\hspace{9mm}\textcolor{DeepBlue}{- ...... }} \\
    \textit{\hspace{9mm}\textcolor{DeepBlue}{- 273.64s $\sim$ 319.18s: Make egg salad (make egg salad)}} \\
    \textit{\hspace{12mm}\textcolor{DeepBlue}{- (273.72s $\sim$ 299.58s: Cut onion (cut onion on kitchen board))}} \\
    \textit{\hspace{12mm}\textcolor{DeepBlue}{- (299.58s $\sim$ 310.41s: Add onion to recipe (add chopped onions to plate))}} \\
    \#\# Output (Specifies the output format and requirements, such as generating five diverse QA pairs in JSON format.) \\
    \midrule
    \textcolor{blue}{\textbf{Constraints and Guidelines}} (Defines rules and limitations to ensure the generated content is clear, specific, and natural, avoiding ambiguity and technical jargon.) \\
    \end{tabular}
\end{tcolorbox}
\caption{Generalized Prompt Template for Video Analysis Tasks. The content highlighted in \textcolor{blue}{blue} changes according to the specific task.}
\label{tab:Generalized Prompt Template for Video Analysis Tasks}
\end{minipage}
\end{table*}

\begin{CJK*}{UTF8}{gbsn}
\begin{table*}[h!]
\centering
\begin{minipage}{0.95\textwidth}
\centering
\begin{tcolorbox}
    \centering
    \small
    \begin{tabular}{>{\raggedright\arraybackslash}p{0.95\textwidth}}
    \textcolor{blue}{\textbf{System Message}} (用于为整个任务设定背景和角色，不仅指明系统应扮演的角色，还对任务性质进行定义。) \\
    \textit{例子：系统消息根据任务粒度的不同分为两个层次：} \\
    \textit{- \textbf{粗粒度} (聚焦事件层面，引导 GPT-4 生成涉及动作序列识别与特定动作定位的问题。为正确回答此类问题，视频大模型需通过时序推理识别关键动作并理解其时间顺序。)} \\
    \textit{- \textbf{细粒度} (旨在引导 GPT-4 生成关于视频中物体出现情况的问题，但设计核心仍聚焦于视频动态动作的理解，而非静态视觉分析。正确回答此类问题要求视频大模型通过时序推理，不仅要识别物体，还需推断其出现顺序及其与视频动作时间轴的关联性。)} \\
    \midrule
    \textcolor{blue}{\textbf{Prompt Structure}} (为任务提供了明确的操作指南，描述了生成提示时需要遵循的核心要求。) \\
    - \textbf{\#\# Instruction} (为任务提供了明确的操作指南，描述了生成提示时需要遵循的核心要求，如对视频内容进行深入分析并设计具有挑战性的问题及对应的回答。) \\
    - \textbf{\#\# Input} (定义了所需输入的数据格式和内容，主要包括视频的基本信息，如视频主要目标及其关键信息（例如特定时间段内发生的重要事件或步骤）。通过规范输入格式，确保模型获得充分的上下文，从而更好地理解视频内容并生成合理的问题和答案。) \\
    \textit{Example:} \\
    \textit{\hspace{3mm}\textcolor{DeepBlue}{- Instructional Video Clip Info (a segment extracted from the original video):}} \\
    \textit{\hspace{6mm}\textcolor{DeepBlue}{- Video Goal: Making pan seared meal}} \\
    \textit{\hspace{6mm}\textcolor{DeepBlue}{- Key Actions Within Specific Time Intervals (divided into steps and substeps, shown by different indents):}} \\
    \textit{\hspace{9mm}\textcolor{DeepBlue}{- 0.0s $\sim$ 27.16s: Add butter to a pan (heat butter)}} \\
    \textit{\hspace{12mm}\textcolor{DeepBlue}{- (0.16s $\sim$ 18.61s: Add oil to a pan (add butter to skillet))}} \\
    \textit{\hspace{12mm}\textcolor{DeepBlue}{- (18.61s $\sim$ 27.2s: Store ingredients in refrigerator or freezer (put butter in refrigerator))}} \\
    \textit{\hspace{9mm}\textcolor{DeepBlue}{- ...... }} \\
    \textit{\hspace{9mm}\textcolor{DeepBlue}{- 273.64s $\sim$ 319.18s: Make egg salad (make egg salad)}} \\
    \textit{\hspace{12mm}\textcolor{DeepBlue}{- (273.72s $\sim$ 299.58s: Cut onion (cut onion on kitchen board))}} \\
    \textit{\hspace{12mm}\textcolor{DeepBlue}{- (299.58s $\sim$ 310.41s: Add onion to recipe (add chopped onions to plate))}} \\
    - \textbf{\#\# Output} (规定了生成结果的格式和要求，通常需要将问题、答案以及相关解释或设计理由以结构化形式输出（例如 JSON 格式）。本章指出，此部分明确了生成结果应具备的标准和简洁性，以保证输出既符合任务需求又便于后续处理和分析。) \\
    \midrule
    \textcolor{blue}{\textbf{Constraints and Guidelines}} (虽然整体描述中可对数据集视角等内容做后续补充，但本章主要提供生成过程中需要遵守的基本规则和限制。这些规则确保生成的内容具有明确性和针对性，避免出现过于宽泛或技术性过强的表述，从而使生成的提示既自然又符合实际应用要求。) \\
    \end{tabular}
\end{tcolorbox}
\caption{Generalized Prompt Template for Video Analysis Tasks. The content highlighted in \textcolor{blue}{blue} changes according to the specific task.}
\label{tab:Generalized Prompt Template for Video Analysis Tasks}
\end{minipage}
\end{table*}

\end{CJK*}

\begin{table*}[h!]
\centering
\begin{minipage}{0.95\textwidth}
\centering
\begin{tcolorbox}
    \centering
    \small
    \begin{tabular}{p{0.95\textwidth}}
    \textcolor{blue}{\textbf{System message}} \\
    You excel in interpreting video footage recorded from a first-person perspective by a pair of smart glasses, where an Augmented Reality (AR) assistant is integrated. Your skill set includes capturing detailed objects in these videos and extracting their related information. Your task is to anticipate potential user questions and provide relevant and helpful responses. In this task, you will play two roles: \\
    \hspace{3mm}- The User: Imagine you're wearing smart glasses, and you want to recall past detailed objects and extract their related info. Ask questions in a vivid and natural manner, based on the detailed objects you observe, as well as the related objects and actions associated with these observed details, as if you were personally experiencing the events unfolding in the video.\\
    \hspace{3mm}- The Augmented Reality (AR) assistant: Respond to the user's questions. Your responses should be relevant and helpful, using the detailed objects you've observed, as well as the information about related objects and actions associated with these observed details. \\
    \midrule
    \textcolor{blue}{\textbf{Prompt}} \\
    \#\# Instruction:\\
        \hspace{3mm}- I'll provide you with text-based video clip info, which is a segment extracted from an instructional video.\\
        \hspace{3mm}- The video clip info includes the video's goal (\texttt{"video goal"}), duration (\texttt{"duration"}), and \texttt{"key action within specific time intervals"}, which are composed of two layers: actions and sub-actions, shown by different indents.\\
        \hspace{3mm}- Thoroughly examine the text video info and formulate ten detailed, Fine-Grained, challenging question-answer pairs. As if you are watching the video, ask an AR assistant. Then, as the AR assistant, provide useful response.\\
    \#\#  Input:\\
        \hspace{3mm}\textcolor{DeepBlue}{- Instructional Video Clip Info(a segment extracted from the original video):}\\
        \hspace{6mm}\textcolor{DeepBlue}{- video goal: Making rice, dough and vegetable dish}\\
        \hspace{6mm}\textcolor{DeepBlue}{- key action within specific time intervals(consisting of two layers: actions and sub-actions, indicated by different idents):}\\
            \hspace{9mm}\textcolor{DeepBlue}{- 1069.34s $\sim$ 1298.59s:  Preheat a pan or pot on the stovetop(preheat pots)}\\
                \hspace{12mm}\textcolor{DeepBlue}{- (1070.09s $\sim$ 1158.54s:  Check and adjust the heat on a wood stove(Add wood to fire))}\\
                \hspace{9mm}\textcolor{DeepBlue}{......} \\
                \hspace{9mm}\textcolor{DeepBlue}{- 1432.4s $\sim$ 1592.63s:  Get ingredients from pantry or shelf(sieve flour into a tray)}\\
                \hspace{12mm}\textcolor{DeepBlue}{- (1452.19s $\sim$ 1550.37s:  Sieve or sift dry ingredients like flour or grains(Sieve flour))}\\
    \#\#  Output:\\
        \hspace{3mm}- Use your extensive experience as much as possible to design ten diverse, Fine-Grained, detailed, challenging questions that fully test the AR assistant's ability to capture the detailed objects and extract the related objects' action after watching this video clip. Below are some question types you can refer, accompanied by design ideas for crafting them:\\
            \hspace{6mm}- DO NOT ASK coarsed questions based on action, but ask detailed questions based on the objects.\\
            \hspace{6mm}- You can ask what object(s) are included in an action(s). To answer the question, the assistant should find the action, and extract the object info.\\
            \hspace{6mm}- You can ask about the detailed attribute info with the objects you found, such as its appearance, color, size, its position, its type, etc. To answer the question, the assistant need to watch the video carefully, and find the object, and recognize its detailed attributes.\\
            \hspace{6mm}- You can ask about the interactions associated with the objects you found, including the related objects and the related actions. To answer the question, the assistant need carefully watch the video, find the object, extract the interaction info of the object. But DO NOT USE the words \texttt{"objects"} and \texttt{"interactions"} in your questions, ask questions naturally and vividly as if you were watching a video. \\
        \hspace{3mm}- Note: the \texttt{"Instructional Video Clip Info"} do not cover all the information in the video. Therefore, DO NOT USE absolute ordinal numbers like \texttt{"first thing"}, \texttt{"first"}, \texttt{"last"}, \texttt{"second"},\texttt{"third"} etc. or other absolute words like \texttt{"only"}, \texttt{"beginning"}, \texttt{"final"}, \texttt{"end"} etc. when asking questions or answering. \\
        \hspace{3mm}- DO NOT MENTION \texttt{"video info"} \texttt{"step"} \texttt{"substep"} \texttt{"subaction"} \texttt{"the individual"} \texttt{"the person"} etc. Pose questions and provide answers in a vivid and natural way, as if you were a user interacting with the AR assistant while doing daily activities, and as if you were the AR assistant responding to the user's inquiries. \\
        \hspace{3mm}- DO NOT include \texttt{"time interval"} \texttt{"timestamp"} or any number realted to the specific time in your answers or questions.\\
        \hspace{3mm}- ONLY ASK specific question that have definite answers. DO NOT ASK any question that cannot be answered confidently with the information from the provided video text info. DO NOT ASK ambiguous questions, DO NOT ASK overly generalized questions that may leave the assitant uncertain about how to respond. \\
        \hspace{3mm}- Focus on the specificity and challenge of each question. Avoid questions that are too broad or too simple.\\
        \hspace{3mm}- Please ensure that the questions you generate are not repetitive and cover as much of the video content as possible.\\
        \hspace{3mm}- Give a consice, brief reference answer to each question, each Answer should be under 20 words. And your answer must be a direct reference to the information provided. DO NOT include anything that wasn't in the original video info.\\
        \hspace{3mm}- Return a list that includes ten question-answer pairs, and for each pair, provide a designed reason that explains the question type and where the answer comes from. Each pair should be in JSON format: \{\texttt{"Question"}: xxx, \texttt{"Answer"}: xxx, \texttt{"Reason"}: xxx\}.\\
    \end{tabular}
\end{tcolorbox}
\caption{Prompt template designed for generating \textbf{Fine-Grained} temporal reasoning questions for videos in the \textbf{Ego4D Goal-Step} dataset. The content highlighted in \textcolor{DeepBlue}{blue} changes according to the specific video.}
    \label{tab:Prompt Template for Ego4D Goal-Step Fine-Grained Q&A Generation}
\end{minipage}
\end{table*}

\begin{table*}[h!]
\centering
\begin{minipage}{0.95\textwidth}
\centering
\begin{tcolorbox}
    \centering
    \small
    \begin{tabular}{p{0.95\textwidth}}
    \textcolor{blue}{\textbf{System message}} \\
    You excel in interpreting video footage recorded from a first-person perspective by a pair of smart glasses, where an Augmented Reality (AR) assistant is integrated. Your skill set includes capturing the actions sequence and Fine-Grained actions of the video. Your task is to anticipate potential user questions and provide relevant and helpful responses. In this task, you will play two roles: \\
    \hspace{3mm}- The User: Imagine you're wearing smart glasses, and you want to recall past actions, actions sequence or predict what will happen next. Ask questions that recall past actions, actions sequence or predict future actions in a vivid and natural manner, as if you were really there.\\
    \hspace{3mm}- The Augmented Reality (AR) assistant: Answer the user's questions. Your answers should be useful and should be based on your understanding of the main actions and action sequences in the video. \\
    \midrule
    \textcolor{blue}{\textbf{Prompt}} \\
    \#\# Instruction:\\
        \hspace{3mm}- I'll provide you with text-based video clip info, which is a segment extracted from an instructional video.\\
        \hspace{3mm}- The video clip info includes the video's scenario (\texttt{"video scenario"}), duration (\texttt{"duration"}), and \texttt{"key action within specific time intervals"}.\\
        \hspace{3mm}- Thoroughly examine the text video info and design ten inferential, challenging question-answer pairs. As if you were watching the video, ask an AR assistant. Then, as the AR assistant, provide useful response.\\
    \#\#  Input:\\
        \hspace{3mm}\textcolor{DeepBlue}{- Instructional Video Clip Info(a segment extracted from the original video):}\\
        \hspace{6mm}\textcolor{DeepBlue}{- video scenario: Covid-19 Rapid Antigen Test}\\
        \hspace{6mm}\textcolor{DeepBlue}{- key action within specific time intervals:}\\
            \hspace{9mm}\textcolor{DeepBlue}{- 0.0s $\sim$ 24.57s: Unbox package(Unbox the COVID-19 test kit box)}\\
            \hspace{9mm}\textcolor{DeepBlue}{- 24.63s $\sim$ 33.78s: Arrange test material(Arrange the COVID-19 test materials on the table)}\\
            \hspace{9mm}\textcolor{DeepBlue}{......} \\
            \hspace{9mm}\textcolor{DeepBlue}{- 237.41s $\sim$ 243.18s: Set the swab into the testing plate(Insert the tip of the COVID-19 collection swab into the COVID-19 test card)}\\
    \#\#  Output:\\
        \hspace{3mm}- Use your extensive experience as much as possible to design ten diverse, challenging, inferential questions that fully test the AR assistant's ability to capture actions sequence and Fine-Grained actions after watching this video clip. Below are some question types you can refer, accompanied by design ideas for crafting them, and feel free to enrich them with your wisdom and experience:\\
            \hspace{6mm}- You can design questions to list the sequence between any selected two or more key actions, but DO NOT directly copy the punctuation marks from the given info when referring to the actions.\\
            \hspace{6mm}- You can just randomly select one of the key action(s) and ask what are the action(s) before and after it to check whether the assistants have watched and understood the whole video. To answer question the assistant must refer to the video and compare different parts of the video.\\
            \hspace{6mm}- You can ask the assistant summarize the key actions between randomly slected two actions.\\
            \hspace{6mm}- Refer naturally to key actions info using words such as\texttt{"before"}, \texttt{"earlier"}, \texttt{"prior to"}, \texttt{"following"}, \texttt{"after"}, \texttt{"subsequently"}, \texttt{"finally"}, etc., Do Not Mention \texttt{"action"} \texttt{"step"} or \texttt{"steps"} when asking questions.\\
        \hspace{3mm}- Note: the \texttt{"Instructional Video Clip Info"} do not cover all the information in the video. Therefore, DO NOT USE absolute ordinal numbers like \texttt{"first thing"}, \texttt{"first"}, \texttt{"last"}, \texttt{"second"},\texttt{"third"} etc. or other absolute words like \texttt{"only"}, \texttt{"beginning"}, \texttt{"final"}, \texttt{"end"} etc. when asking questions or answering. \\
        \hspace{3mm}- DO NOT MENTION \texttt{"video info"}, \texttt{"step"}, \texttt{"substep"}, \texttt{"subaction"}, \texttt{"the individual"}, \texttt{"the person"} etc. Pose questions and provide answers in a vivid and natural way, as if you were a user interacting with the AR assistant while doing daily activities, and as if you were the AR assistant responding to the user's inquiries. \\
        \hspace{3mm}- DO NOT include \texttt{"time interval"}, \texttt{"timestamp"} or any number related to the specific time in your answers or questions.\\
        \hspace{3mm}- ONLY ASK specific questions that have definite answers. DO NOT ASK any question that cannot be answered confidently with the information from the provided video text info. DO NOT ASK ambiguous questions, DO NOT ASK overly generalized questions that may leave the assistant uncertain about how to respond. \\
        \hspace{3mm}- Focus on the specificity and challenge of each question. Avoid questions that are too broad or too simple.\\
        \hspace{3mm}- Please ensure that the questions you generate are not repetitive and cover as much of the video content as possible.\\
        \hspace{3mm}- Give a concise, brief reference answer to each question, each Answer should be under 20 words. And your answer must be a direct reference to the information provided. DO NOT include anything that wasn't in the original video info.\\
        \hspace{3mm}- Return a list that includes ten question-answer pairs, and for each pair, provide a designed reason that explains the question type and where the answer comes from. Each pair should be in JSON format: \{\texttt{"Question"}: xxx, \texttt{"Answer"}: xxx, \texttt{"Reason"}: xxx\}.\\
    \end{tabular}
\end{tcolorbox}
\caption{Prompt template designed for generating \textbf{Coarse-Grained} temporal reasoning questions for videos in the \textbf{Ego-Exo4D} dataset. The content highlighted in \textcolor{DeepBlue}{blue} changes according to the specific video.}
\label{tab:Prompt Template for Ego-Exo4D Coarse-Grained Q&A Generation}
\end{minipage}
\end{table*}

\begin{table*}[h!]
\centering
\begin{minipage}{0.95\textwidth}
\centering
\begin{tcolorbox}
    \centering
    \small
    \begin{tabular}{p{0.95\textwidth}}
    \textcolor{blue}{\textbf{System message}} \\
    You excel in interpreting video footage recorded from a first-person perspective by a pair of smart glasses, where an Augmented Reality (AR) assistant is integrated. Your skill set includes capturing detailed objects in these videos and extracting their related information. Your task is to anticipate potential user questions and provide relevant and helpful responses. In this task, you will play two roles: \\
    \hspace{3mm}- The User: Imagine you're wearing smart glasses, and you want to recall past details objects and extract their related info. Ask questions in a vivid and natural manner, based on the detailed objects you observe, as well as the related objects and actions associated with these observed details, as if you were personally experiencing the events unfolding in the video.\\
    \hspace{3mm}- The Augmented Reality (AR) assistant: Respond to the user's questions. Your responses should be relevant and helpful, using the detailed objects you've observed, as well as the information about related objects and actions associated with these observed details. \\
    \midrule
    \textcolor{blue}{\textbf{Prompt}} \\
    \#\# Instruction:\\
        \hspace{3mm}- I'll provide you with text-based video clip info, which is a segment extracted from an instructional video.\\
        \hspace{3mm}- The video clip info includes the video's scenario (\texttt{"video scenario"}), duration (\texttt{"duration"}), and \texttt{"key action within specific time intervals"}.\\
        \hspace{3mm}- Thoroughly examine the text video info and formulate ten detailed, Fine-Grained, challenging question-answer pairs. As if you are watching the video, ask an AR assistant. Then, as the AR assistant, provide useful response.\\
    \#\#  Input:\\
        \hspace{3mm}\textcolor{DeepBlue}{- Instructional Video Clip Info(a segment extracted from the original video):}\\
        \hspace{6mm}\textcolor{DeepBlue}{- video scenario: Covid-19 Rapid Antigen Test}\\
        \hspace{6mm}\textcolor{DeepBlue}{- key action within specific time intervals:}\\
            \hspace{9mm}\textcolor{DeepBlue}{- 0.0s $\sim$ 24.57s: Unbox package(Unbox the COVID-19 test kit box)}\\
            \hspace{9mm}\textcolor{DeepBlue} {.......}\\
            \hspace{9mm}\textcolor{DeepBlue}{- 237.41s $\sim$ 243.18s: Set the swab into the testing plate(Insert the tip of the COVID-19 collection swab into the COVID-19 test card)}\\
    \#\#  Output:\\
        \hspace{3mm}- Use your extensive experience as much as possible to design ten diverse, Fine-Grained, detailed, challenging questions that fully test the AR assistant's ability to capture the detailed objects and extract the related objects' action after watching this video clip. Below are some question types you can refer, accompanied by design ideas for crafting them:\\
            \hspace{6mm}- DO NOT ASK coarsed questions based on action, but ask detailed questions based on the objects.\\
            \hspace{6mm}- You can ask what object(s) are included in an action(s). To answer the question, the assistant should find the action, and extract the object info.\\
            \hspace{6mm}- You can ask about the detailed attribute info with the objects you found, such as its appearance, color, size, its position, its type, etc. To answer the question, the assistant need to watch the video carefully, and find the object, and recognize its detailed attributes.\\
            \hspace{6mm}- You can ask about the interactions associated with the objects you found, including the related objects and the related actions. To answer the question, the assistant need carefully watch the video, find the object, extract the interaction info of the object. But DO NOT USE the words \texttt{"objects"} and \texttt{"interactions"} in your questions, ask questions naturally and vividly as if you were watching a video. \\
        \hspace{3mm}- Note: the \texttt{"Instructional Video Clip Info"} do not cover all the information in the video. Therefore, DO NOT USE absolute ordinal numbers like \texttt{"first thing"}, \texttt{"first"}, \texttt{"last"}, \texttt{"second"},\texttt{"third"} etc. or other absolute words like \texttt{"only"}, \texttt{"beginning"}, \texttt{"final"}, \texttt{"end"} etc. when asking questions or answering. \\
        \hspace{3mm}- DO NOT MENTION \texttt{"video info"}, \texttt{"step"}, \texttt{"substep"}, \texttt{"subaction"}, \texttt{"the individual"}, \texttt{"the person"} etc. Pose questions and provide answers in a vivid and natural way, as if you were a user interacting with the AR assistant while doing daily activities, and as if you were the AR assistant responding to the user's inquiries. \\
        \hspace{3mm}- DO NOT include \texttt{"time interval"}, \texttt{"timestamp"} or any number related to the specific time in your answers or questions.\\
        \hspace{3mm}- ONLY ASK specific questions that have definite answers. DO NOT ASK any question that cannot be answered confidently with the information from the provided video text info. DO NOT ASK ambiguous questions, DO NOT ASK overly generalized questions that may leave the assistant uncertain about how to respond. \\
        \hspace{3mm}- Focus on the specificity and challenge of each question. Avoid questions that are too broad or too simple.\\
        \hspace{3mm}- Please ensure that the questions you generate are not repetitive and cover as much of the video content as possible.\\
        \hspace{3mm}- Give a concise, brief reference answer to each question, each Answer should be under 20 words. And your answer must be a direct reference to the information provided. DO NOT include anything that wasn't in the original video info.\\
        \hspace{3mm}- Return a list that includes ten question-answer pairs, and for each pair, provide a designed reason that explains the question type and where the answer comes from. Each pair should be in JSON format: \{\texttt{"Question"}: xxx, \texttt{"Answer"}: xxx, \texttt{"Reason"}: xxx\}.\\
    \end{tabular}
\end{tcolorbox}
\caption{Prompt template designed for generating \textbf{Fine-Grained} temporal reasoning questions for videos in the \textbf{Ego-Exo4D} dataset. The content highlighted in \textcolor{DeepBlue}{blue} changes according to the specific video.}
    \label{tab:Prompt Template for Ego-Exo4D Fine-Grained Q&A Generation}
\end{minipage}
\end{table*}

\begin{table*}[t!]\centering
\begin{minipage}{0.95\textwidth}
\centering
\begin{tcolorbox} 
    \centering
      \small
    \begin{tabular}{p{0.95\textwidth}}
   \textcolor{blue}{\textbf{System message}} \\
You are tasked to classify questions based on the type of video understanding task they relate to. As an experienced expert in video understanding, you are expected to apply your deep knowledge to accurately categorize each question.\\
    \midrule
   \textcolor{blue}{\textbf{Prompt}} \\
\#\# Instruction:\\
    \hspace{3mm}- Please carefully read the provided question and classify it by selecting the most appropriate category from the list below. Use the examples as a guide to understand the types of questions typical for each category. If the question does not fit any of the listed categories, please classify it under ``Others'' and \textbf{specify a unique task name} that best describes the question's type.\\
    \hspace{6mm}- Step Sequencing:\\
        \hspace{9mm}Example: ``What does one do first, mash the potatoes or add them with onions, cheese, salt, and pepper into a bowl to mix?''\\
    \hspace{6mm}- Specific Step Recognition:\\
        \hspace{9mm}Example: ``Which step/action involves checking the temperature of the oven?''\\
    \hspace{6mm}- Past Step Recall:\\
        \hspace{9mm}Example: ``What step/action comes just before adding the eggs to the mix?''\\
    \hspace{6mm}- Future Step Prediction:\\
        \hspace{9mm}Example: ``What follows after kneading the dough?''\\
    \hspace{6mm}- Intermediate Steps Recognition:\\
        \hspace{9mm}Example: ``What steps/actions occur between preheating the oven and placing the cake in it?''\\
    \hspace{6mm}- Object Attribute Recognition:\\
        \hspace{9mm}Example: ``What color is the car that arrives second in the video?''\\
    \hspace{6mm}- Object Existence in Steps:\\
        \hspace{9mm}Example: ``During which step/action is the mixer first used?''\\
    \hspace{6mm}- Object Interaction Recognition:\\
        \hspace{9mm}Example: ``How does the chef use the knife in the step/action involving vegetables?''\\
    \hspace{6mm}- Others:\\
\#\#  Input:\\
\hspace{3mm}\textcolor{DeepBlue}{- Question: What is the process that takes place after cleaning out the face?}\\
\#\#  Output:\\
    \hspace{3mm}- Please structure all of your response in the following JSON format: \{\texttt{"Question\_type"}: xxx, \texttt{"Reason"}: xxx\}. If you classify a question as ``Others'', \textbf{include a specific task name} that accurately reflects the question's type, for example, ``Others(specific task name)''. Ensure that all details are included within the JSON structure and there are no responses outside of it. Ensure the output is valid JSON as it will be parsed using \texttt{json.loads()} in Python.\\
    \hspace{3mm}- Valid response with recognized task type: future step prediction\\
    \end{tabular}
\end{tcolorbox}
\caption{Prompt Template for Question Categorization. The content highlighted in \textcolor{DeepBlue}{blue} changes according to the specific Q\&A.}
\label{tab:Prompt Template for Question Categorization}
\end{minipage}
\end{table*}

\begin{table*}[t!]\centering
\begin{minipage}{0.95\textwidth}
\centering
\begin{tcolorbox} 
    \centering
      \small
    \begin{tabular}{p{0.95\textwidth}}
    \textcolor{blue}{\textbf{System message}} \\
    \hspace{3mm}-\\
    \midrule
   \textcolor{blue}{\textbf{Prompt}} \\
\#\# Input:\\
\hspace{3mm}\textcolor{DeepBlue}{- Question: What happens to the noodles after being boiled and where does it join another ingredient?}\\
\#\# Output:\\
    \end{tabular}
\end{tcolorbox}
\caption{Prompt Template for Video Question Answering Without Video and Blind Question Filtering. The content highlighted in \textcolor{DeepBlue}{blue} changes according to the specific Q\&A.}
\label{tab:Prompt Template for Video Question Answering Without Video and Blind Question Filtering}
\end{minipage}
\end{table*}

\begin{table*}[t!]\centering
\begin{minipage}{0.95\textwidth}
\centering
\begin{tcolorbox} 
    \centering
      \small
    \begin{tabular}{p{0.95\textwidth}}
   \textcolor{blue}{\textbf{System message}} \\
You are an experienced video question designer. Please help me identify any poorly-designed video questions.\\
    \midrule
   \textcolor{blue}{\textbf{Prompt}} \\
\#\# Instruction:\\
    \hspace{3mm}- The user will provide the 'Video\_info' and a 'Question'. The 'Video\_info' describes the video's content.\\
    \hspace{3mm}- Evaluate the question based on the following criteria:\\
    \hspace{6mm}- The question should be related to the 'Video\_info'.\\
    \hspace{6mm}- The question should not ask for information that the 'Video\_info' does not provide. As the 'Video\_info' may not encompass all the details of the video, questions should only be based on the information provided in the 'Video\_info'.\\
    \hspace{6mm}- The question should not be implicit, such as which action lasted the longest and other extreme value questions, unless such information is explicitly mentioned in the detailed 'action' description.\\
    \hspace{6mm}- The question should not require calculations based on the time intervals, such as calculating the duration of an action based on the time intervals. Because the 'Video\_info' may not encompass all the details of the video, it is not correct to calculate the time solely based on the time interval.\\
    \hspace{6mm}- The question should not be subjective or ambiguously phrased.\\
    \hspace{6mm}- The question should not contain meaningless assumptions or abnormal conditions. All questions should be practical and logical to ensure that we can derive valid and useful answers.\\
    \hspace{3mm}- If the 'Question' fails to meet any of the above criteria, it should be considered a poorly-designed question.\\
\#\# Input:\\
\hspace{3mm}\textcolor{DeepBlue}{- Video\_info:}\\
\hspace{6mm}\textcolor{DeepBlue}{- video theme: Make Coconut Yogurt}\\
\hspace{6mm}\textcolor{DeepBlue}{- duration: 194.08s}\\
\hspace{6mm}\textcolor{DeepBlue}{- key action within specific time intervals:}\\
\hspace{9mm}\textcolor{DeepBlue}{- 85s $\sim$ 95s : take coconut milk in bowl}\\
\hspace{9mm}\textcolor{DeepBlue}{- 95s $\sim$ 100s : string well with vinegar}\\
\hspace{9mm}\textcolor{DeepBlue}{- 100s $\sim$ 106s : pour the remaining coconut liquid into that}\\
\hspace{9mm}\textcolor{DeepBlue}{- 106s $\sim$ 117s : pour water into yogurt maker}\\
\hspace{9mm}\textcolor{DeepBlue}{- 117s $\sim$ 141s : put the liquid coconut mix into that}\\
\hspace{9mm}\textcolor{DeepBlue}{- 141s $\sim$ 148s : take the prepared yogurt into glass}\\
\hspace{9mm}\textcolor{DeepBlue}{- 148s $\sim$ 151s : mix it with fruits}\\
\hspace{3mm}\textcolor{DeepBlue}{- Question: What happens after the liquid coconut mix is placed into the yogurt maker?}\\
\#\# Output:\\
    \hspace{3mm}- If the question is poorly-designed, please return the judgement with 'Yes', otherwise 'No'. And give the 'Reason' for the judgement.\\
    \hspace{3mm}- Please structure all of your response in the following JSON format: \{\texttt{"Judgement"}: xxx, \texttt{"Reason"}: xxx\}. Ensure that all details are included within the JSON structure and there are no responses outside of it. Ensure the output is valid JSON as it will be parsed using \texttt{json.loads()} in Python.\\
    \end{tabular}
\end{tcolorbox}
\caption{Prompt template for Hallucination Question Filtering. The content highlighted in \textcolor{DeepBlue}{blue} changes according to the specific Q\&A.}
\label{tab:Prompt template for Hallucination Question Filtering}
\end{minipage}
\end{table*}

\begin{table*}[t!]\centering
\begin{minipage}{0.95\textwidth}
\centering
\begin{tcolorbox} 
    \centering
      \small
    \begin{tabular}{p{0.95\textwidth}}
   \textcolor{blue}{\textbf{System message}} \\
You are an intelligent chatbot. Your task is to answer video questions based on the provided text video info.\\
    \midrule
   \textcolor{blue}{\textbf{Prompt}} \\
\#\# Instruction:\\
    \hspace{3mm}- The user will input a 'Question' and 'Video\_info', and you need to carefully read the 'Video\_info' and answer the 'Question'.\\
    \hspace{3mm}- Answer as if you have watched the video.\\
\#\# Input:\\
\hspace{3mm}\textcolor{DeepBlue}{- Question: Summarize what was done between the time the tissue holder on the center was removed and when the tissue was taken out at the front.}\\
\hspace{3mm}\textcolor{DeepBlue}{- Video\_info:}\\
\hspace{6mm}\textcolor{DeepBlue}{- video theme: Make a Toilet Paper Roll Basket}\\
\hspace{6mm}\textcolor{DeepBlue}{- duration: 113.7s}\\
\hspace{6mm}\textcolor{DeepBlue}{- key action within specific time intervals:}\\
\hspace{9mm}\textcolor{DeepBlue}{- 6s $\sim$ 28s : take wanted materials}\\
\hspace{9mm}\textcolor{DeepBlue}{- 28s $\sim$ 48s : open a big box}\\
\hspace{9mm}\textcolor{DeepBlue}{- 48s $\sim$ 55s : remove tissue holder on center}\\
\hspace{9mm}\textcolor{DeepBlue}{- 55s $\sim$ 63s : wrap the center of tissue}\\
\hspace{9mm}\textcolor{DeepBlue}{- 63s $\sim$ 71s : put tissue on open box}\\
\hspace{9mm}\textcolor{DeepBlue}{- 71s $\sim$ 73s : cover side using tape}\\
\hspace{9mm}\textcolor{DeepBlue}{- 73s $\sim$ 78s : take out tissue on front}\\
\#\# Output:\\
    \hspace{3mm}- Give a concise, brief reference Answer to each question, each Answer should be under 20 words. And your answer must be a direct reference to the information provided.\\
    \hspace{3mm}- Also, give the reason, explaining how you arrived at the answer.\\
    \hspace{3mm}- Please structure all of your response in the following JSON format: \{\texttt{"Answer"}: xxx, \texttt{"Reason"}: xxx\}. Ensure that all details are included within the JSON structure and there are no responses outside of it. Ensure the output is valid JSON as it will be parsed using \texttt{json.loads()} in Python.\\
    \end{tabular}
\end{tcolorbox}
\caption{Prompt Template for Video Question Answering With Video and Tricky Question Filtering. The content highlighted in \textcolor{DeepBlue}{blue} changes according to the specific Q\&A.}
\label{tab:Prompt Template for Video Question Answering With Video and Tricky Question Filtering}
\end{minipage}
\end{table*}

\begin{table*}[t!]\centering
\begin{minipage}{0.95\textwidth}
\centering
\begin{tcolorbox} 
    \centering
      \small
    \begin{tabular}{p{0.95\textwidth}}
   \textcolor{blue}{\textbf{System message}} \\
You are an intelligent and meticulous assistant. Your task is to identify vague, ambiguous, and incomplete answers.\\
    \midrule
   \textcolor{blue}{\textbf{Prompt}} \\
\#\# Instruction:\\
    \hspace{3mm}- Carefully read the 'Video\_info' and 'Question' provided by the user before evaluating the 'Answer'.\\
    \hspace{3mm}- The 'Video\_info' describes the video's content, and the 'Question' is based on the 'Video\_info'. The 'Answer' is a response to the 'Question', but it may be vague, ambiguous, or incomplete.\\
    \hspace{3mm}- Evaluate the 'Answer' based on the following criteria:\\
    \hspace{6mm}- Specificity: The answer should be specific and detailed, not vague. It should directly address the specifics of the question, while keeping the word count within 20 words.\\
    \hspace{6mm}- Completeness: The answer should provide a complete response to the question. It should cover all aspects of the question and not leave any part unanswered.\\
    \hspace{6mm}- Coverage: If there are multiple correct answers to the question based on 'Video\_info' and the 'Question', the answer should include all possible correct responses.\\
    \hspace{3mm}- If the 'Answer' fails to meet any of the above criteria, it should be considered a bad answer.\\
\#\# Input:\\
\hspace{3mm}\textcolor{DeepBlue}{- video\_info:}\\
\hspace{6mm}\textcolor{DeepBlue}{- video theme: Tie a Prusik Knot}\\
\hspace{6mm}\textcolor{DeepBlue}{- duration: 129.26s}\\
\hspace{6mm}\textcolor{DeepBlue}{- key action within specific time intervals:}\\
\hspace{9mm}\textcolor{DeepBlue}{- 9s $\sim$ 16s : Get cordage}\\
\hspace{9mm}\textcolor{DeepBlue}{- 16s $\sim$ 22s : Fold it in half}\\
\hspace{9mm}\textcolor{DeepBlue}{- 22s $\sim$ 27s : Put it over the line}\\
\hspace{9mm}\textcolor{DeepBlue}{- 27s $\sim$ 35s : Push it through}\\
\hspace{9mm}\textcolor{DeepBlue}{- 35s $\sim$ 44s : Repeat the process several times}\\
\hspace{9mm}\textcolor{DeepBlue}{- 44s $\sim$ 54s : Keep the pair apart}\\
\hspace{9mm}\textcolor{DeepBlue}{- 54s $\sim$ 61s : Pull it through}\\
\hspace{3mm}\textcolor{DeepBlue}{- Question: What action is repeatedly done?}\\
\hspace{3mm}\textcolor{DeepBlue}{- Answer: The process of pushing it through is repeatedly done.}\\
\#\# Output:\\
    \hspace{3mm}- If the answer is bad, please return the judgement with 'Yes', otherwise 'No'. And give the 'Reason' for the judgement.\\
    \hspace{3mm}- Please structure all of your response in the following JSON format: \{\texttt{"Judgement"}: xxx, \texttt{"Reason"}: xxx\}. Ensure that all details are included within the JSON structure and there are no responses outside of it. Ensure the output is valid JSON as it will be parsed using \texttt{json.loads()} in Python.\\
    \end{tabular}
\end{tcolorbox}
\caption{Prompt Template for Unspecific and Incomplete Answers Filtering. The content highlighted in \textcolor{DeepBlue}{blue} changes according to the specific Q\&A.}
\label{tab:Prompt Template for Unspecific and Incomplete Answers Filtering}
\end{minipage}
\end{table*}

\begin{table*}[t!]\centering
\begin{minipage}{0.95\textwidth}
\centering
\begin{tcolorbox} 
    \centering
      \small
    \begin{tabular}{p{0.95\textwidth}}
   \textcolor{blue}{\textbf{System message}} \\
You are an intelligent and careful teacher. Your task is to carefully check whether the student's answer is right.\\
    \midrule
   \textcolor{blue}{\textbf{Prompt}} \\
\#\# Instruction:\\
    \hspace{3mm}- The student should answer the question directly, not provide a range of possible answers.\\
    \hspace{3mm}- The student's answer should be closely related to the question, not covering all possible information related to the topic.\\
    \hspace{3mm}- Focus on the factual consistency between the student's answer and the correct answer. The student's answer should not contain any misinterpretations or misinformation.\\
    \hspace{3mm}- The student's answer must be factually accurate and align with the correct answer.\\
    \hspace{3mm}- Consider synonyms or paraphrases as valid matches.\\
\#\# Input:\\
\hspace{3mm}\textcolor{DeepBlue}{- Question: Summarize what was done between the time the tissue holder on the center was removed and when the tissue was taken out at the front.}\\
\hspace{3mm}\textcolor{DeepBlue}{- Correct Answer: After removing the tissue holder, the center of the tissue was wrapped, then put on the open box, the side was covered using tape before finally taking out the tissue on the front.}\\
\hspace{3mm}\textcolor{DeepBlue}{- Student's Answer: The tissue holder in the center was removed, then the center of the tissue was wrapped, it was put on an open box, the side was covered using tape, and then the tissue was taken out at the front.}\\
\#\# Output:\\
    \hspace{3mm}- If the student's answer is correct, please return the judgement with 'Yes', otherwise 'No'. And give the 'Reason' for the judgement.\\
    \hspace{3mm}- Please structure all of your response in the following JSON format: \{\texttt{"Judgement"}: xxx, \texttt{"Reason"}: xxx\}. Ensure that all details are included within the JSON structure and there are no responses outside of it. Ensure the output is valid JSON as it will be parsed using \texttt{json.loads()} in Python.\\
    \end{tabular}
\end{tcolorbox}
\caption{Prompt Template for Checking Consistency Between Predicted and Correct Answers. The content highlighted in \textcolor{DeepBlue}{blue} changes according to the specific Q\&A.}
\label{tab:Prompt Template for Checking Consistency Between Predicted and Correct Answers}
\end{minipage}
\end{table*}


\begin{table*}[t!]\centering
\begin{minipage}{0.95\textwidth}
\centering
\begin{tcolorbox} 
    \centering
      \small
    \begin{tabular}{p{0.95\textwidth}}
   \textcolor{blue}{\textbf{System message}} \\
You are an experienced teacher specializing in creating challenging single-choice questions (SCQs). You will be provided with video text info, a question, and a reference answer related to the video. Your task is to design a single-choice question (SCQ) based on this information.\\
    \midrule
   \textcolor{blue}{\textbf{Prompt}} \\
\#\# Instruction:\\
    \hspace{3mm}- 1. Read the transcript to understand the context thoroughly.\\
    \hspace{3mm}- 2. Review the given question and the reference answer carefully.\\
    \hspace{3mm}- 3. Generate six additional incorrect answer choices. These choices should be:\\
        \hspace{6mm}- Relevance: Ensure all wrong answers are directly related to the main topic or actions discussed in the video or question. Avoid introducing unrelated elements.\\
        \hspace{6mm}- Plausible Alternatives: Design wrong answers that might seem correct by using elements from the video or those commonly associated with the topic. This requires viewers to think critically.\\
        \hspace{6mm}- Subtle Misinformation: Introduce minor errors in the wrong answers, such as slight misinterpretations or incorrect details that are closely related to the correct information. Verify each wrong answer against the video and the question to ensure they remain incorrect.\\
        \hspace{6mm}- Length Consistency: Maintain a length similar to the correct answer to prevent elimination based on length discrepancies.\\
        \hspace{6mm}- Complexity: Avoid overly simplistic or generic wrong answers. They should reflect a nuanced misunderstanding of the content to enhance the challenge.\\
    \hspace{3mm}- 4. Ensure that the answer choices are challenging yet fair, aimed at testing the student's comprehension and critical thinking skills.\\
\#\#  Input:\\
\hspace{3mm}\textcolor{DeepBlue}{- Video\_info:}\\
\hspace{6mm}\textcolor{DeepBlue}{- video theme: Make a Plaster Mask}\\
\hspace{6mm}\textcolor{DeepBlue}{- duration: 103.0s}\\
\hspace{6mm}\textcolor{DeepBlue}{- key action within specific time intervals:}\\
\hspace{9mm}\textcolor{DeepBlue}{- 35s $\sim$ 48s : clean out the face}\\
\hspace{9mm}\textcolor{DeepBlue}{- 48s $\sim$ 60s : apply tissue using water on face}\\
\hspace{9mm}\textcolor{DeepBlue}{- 60s $\sim$ 72s : apply it for full face}\\
\hspace{9mm}\textcolor{DeepBlue}{- 72s $\sim$ 77s : put it under the neck}\\
\hspace{9mm}\textcolor{DeepBlue}{- 77s $\sim$ 79s : dry it out}\\
\hspace{3mm}\textcolor{DeepBlue}{- Question: What is the process that takes place after cleaning out the face?}\\
\hspace{3mm}\textcolor{DeepBlue}{- Answer: After cleaning the face, tissue is applied on the face using water.}\\
\#\#  Output:\\
    \hspace{3mm}- Please structure all of your response in the following JSON format: \{\texttt{"Wrong\_answer1"}: xxx, \texttt{"Wrong\_answer2"}:xxx, \texttt{"Wrong\_answer3"}:xxx, \texttt{"Wrong\_answer4"}:xxx, \texttt{"Wrong\_answer5"}:xxx, \texttt{"Wrong\_answer6"}:xxx\}. Ensure that all details are included within the JSON structure and there are no responses outside of it. Ensure the output is valid JSON as it will be parsed using \texttt{json.loads()} in Python.\\
    \end{tabular}
\end{tcolorbox}
\caption{Prompt Template for 6 Wrong Answers Generation. The content highlighted in \textcolor{DeepBlue}{blue} changes according to the specific Q\&A.}
\label{tab:Prompt Template for 6 Wrong Answers Generation}
\end{minipage}
\end{table*}

\begin{table*}[t!]\centering
\begin{minipage}{0.95\textwidth}
\centering
\begin{tcolorbox} 
    \centering
      \small
    \begin{tabular}{p{0.95\textwidth}}
   \textcolor{blue}{\textbf{System message}} \\
You are an experienced teacher specializing in creating challenging single-choice questions (SCQs). You will be provided with video text info, a question, a reference answer, and six potential wrong answers. Your task is to select the four most appropriate wrong answers based on the provided information and formulate a single-choice question (SCQ).\\
    \midrule
   \textcolor{blue}{\textbf{Prompt}} \\
\#\# Instruction:\\
    \hspace{3mm}- 1.Read the provided video text info thoroughly to understand the context.\\
    \hspace{3mm}- 2.Review the given question and the reference answer carefully.\\
    \hspace{3mm}- 3.Evaluate the all potential wrong answers. Select the four most suitable wrong answers by considering the following criteria:\\
    \hspace{6mm}- Plausibility: The wrong answers should be contextually relevant and plausible at a glance, but clearly incorrect upon closer inspection\\
    \hspace{6mm}- Diversity: The wrong answers should be distinct from one another and the correct answer, avoiding overly similar choices\\
    \hspace{6mm}- Length Consistency: The wrong answers should be of similar length to the reference answer to avoid easy elimination based on length alone.\\
    \hspace{6mm}- Contextual Relevance: The wrong answers should be relevant to the video’s content but not correct upon detailed understanding.\\
    \hspace{6mm}- Uniqueness: Ensure that the selected wrong answers are not repetitive and each one is distinct.\\
\#\# Input:\\
\hspace{3mm}\textcolor{DeepBlue}{- Video\_info:}\\
\hspace{6mm}\textcolor{DeepBlue}{- video theme: Make a Plaster Mask}\\
\hspace{6mm}\textcolor{DeepBlue}{- duration: 103.0s}\\
\hspace{6mm}\textcolor{DeepBlue}{- key action within specific time intervals:}\\
\hspace{9mm}\textcolor{DeepBlue}{- 35s $\sim$ 48s : clean out the face}\\
\hspace{9mm}\textcolor{DeepBlue}{- 48s $\sim$ 60s : apply tissue using water on face}\\
\hspace{9mm}\textcolor{DeepBlue}{- 60s $\sim$ 72s : apply it for full face}\\
\hspace{9mm}\textcolor{DeepBlue}{- 72s $\sim$ 77s : put it under the neck}\\
\hspace{9mm}\textcolor{DeepBlue}{- 77s $\sim$ 79s : dry it out}\\
\hspace{3mm}\textcolor{DeepBlue}{- Question: What is the process that takes place after cleaning out the face?}\\
\hspace{3mm}\textcolor{DeepBlue}{- Answer: After cleaning the face, tissue is applied on the face using water.}\\
\hspace{3mm}\textcolor{DeepBlue}{- Wrong\_answer\_list:}\\
\hspace{6mm}\textcolor{DeepBlue}{- Wrong\_answer: After cleaning the face, it is dried out.}\\
\hspace{6mm}\textcolor{DeepBlue}{- Wrong\_answer: After cleaning the face, another layer of cleaning is applied.}\\
\hspace{6mm}\textcolor{DeepBlue}{- Wrong\_answer: After cleaning the face, the mask is placed under the neck.}\\
\hspace{6mm}\textcolor{DeepBlue}{- Wrong\_answer: After cleaning the face, the mask is fully finished.}\\
\hspace{6mm}\textcolor{DeepBlue}{- Wrong\_answer: After cleaning the face, the mask is immediately used.}\\
\hspace{6mm}\textcolor{DeepBlue}{- Wrong\_answer: After cleaning the face, a brush is used to smooth the mask.}\\
\#\# Output:\\
    \hspace{3mm}- Please structure all of your response in the following JSON format: \{\texttt{"Wrong\_answer1"}: xxx, \texttt{"Wrong\_answer2"}:xxx, \texttt{"Wrong\_answer3"}:xxx, \texttt{"Wrong\_answer4"}:xxx\}. Ensure that all details are included within the JSON structure and there are no responses outside of it. Ensure the output is valid JSON as it will be parsed using \texttt{json.loads()} in Python.\\
    \end{tabular}
\end{tcolorbox}
\caption{Prompt Template for Picking Up 4 Wrong Answers. The content highlighted in \textcolor{DeepBlue}{blue} changes according to the specific Q\&A.}
\label{tab:Prompt Template for Picking Up 4 Wrong Answers}
\end{minipage}
\end{table*}

\begin{table*}[t!]\centering
\begin{minipage}{0.95\textwidth}
\centering
\begin{tcolorbox} 
    \centering
      \small
    \begin{tabular}{p{0.95\textwidth}}
   \textcolor{blue}{\textbf{System message}} \\
You are tasked with helping users choose the best option that accurately answers a question, without the aid of video content.\\
    \midrule
   \textcolor{blue}{\textbf{Prompt}} \\
\#\# Instruction:\\
    \hspace{3mm}- To select the most appropriate option from a set of choices based solely on the question, follow these steps:\\
    \hspace{6mm}- 1. Read the Question: Understand the specific question being asked, focusing on the key details and what is required for a comprehensive answer.\\
    \hspace{6mm}- 2. Review the Options: Carefully examine each option provided. Consider how each one could potentially answer the question based on logical reasoning and available information.\\
    \hspace{6mm}- 3. Apply Critical Thinking:\\
        \hspace{9mm}- Relevance: Determine which option best aligns with the key points of the question.\\
        \hspace{9mm}- Plausibility: Assess whether the option is plausible and makes logical sense in response to the question.\\
        \hspace{9mm}- Completeness: Check if the option addresses all aspects of the question without omitting important details.\\
    \hspace{6mm}- 4. Make a Decision: Based on your analysis, choose the option that most accurately and completely answers the question.\\
    \hspace{6mm}- 5. Justify Your Choice: Be prepared to explain why this option is the best choice, citing reasons based on the logic and completeness of how the option addresses the question.\\
\#\#  Input:\\
\hspace{3mm}\textcolor{DeepBlue}{- Question: What is the process that takes place after cleaning out the face?}\\
\hspace{3mm}\textcolor{DeepBlue}{- Options:}\\
\hspace{6mm}\textcolor{DeepBlue}{A. Just after cleaning the face, the mask is put on the full face.}\\
\hspace{6mm}\textcolor{DeepBlue}{B. After cleaning the face, the tissue is applied under the neck.}\\
\hspace{6mm}\textcolor{DeepBlue}{C. After cleaning the face, the mask is immediately dried out.}\\
\hspace{6mm}\textcolor{DeepBlue}{D. After cleaning the face, a layer of dry tissue is applied on the face.}\\
\hspace{6mm}\textcolor{DeepBlue}{E. After cleaning the face, tissue is applied on the face using water.}\\
\#\#  Output:\\
    \hspace{3mm}- After completing the analysis, state your best option, including both the option letter and its content, and provide a detailed reason for your choice.\\
    \hspace{3mm}- Please structure all of your response in the following JSON format: \{\texttt{"Best\_option"}: xxx, \texttt{"Reason"}: xxx\}. Ensure that the response is structured correctly in JSON format, as it will be parsed using \texttt{json.loads()} in Python.\\
    \end{tabular}
\end{tcolorbox}
\caption{Prompt Template for Choosing The Best Option Without Video and Ensuring Options' Incorrectness. The content highlighted in \textcolor{DeepBlue}{blue} changes according to the specific Q\&A.}
\label{tab:Prompt Template for Choosing The Best Option Without Video and Ensuring Options' Incorrectness}
\end{minipage}
\end{table*}

\begin{table*}[t!]\centering
\begin{minipage}{0.95\textwidth}
\centering
\begin{tcolorbox} 
    \centering
      \small
    \begin{tabular}{p{0.95\textwidth}}
   \textcolor{blue}{\textbf{System message}} \\
You are tasked with helping students choose the best option that accurately answers a question based on the corresponding video information.\\
    \midrule
   \textcolor{blue}{\textbf{Prompt}} \\
\#\# Instruction:\\
    \hspace{3mm}- 1. Review the Video Information: Carefully read and understand the details provided about the video to grasp the context and key content that are relevant to the question.\\
    \hspace{3mm}- 2. Read the Question: Clearly understand the specific question being asked, noting any key details that are crucial for identifying the correct answer.\\
    \hspace{3mm}- 3. Review the Options: Examine each option provided. Consider how each one relates to the video information and the question asked.\\
    \hspace{3mm}- 4. Apply Critical Thinking:\\
    \hspace{6mm}- Relevance: Determine which option best aligns with the key points and details described in the video information.\\
    \hspace{6mm}- Accuracy: Assess whether the option correctly reflects the information and context provided in the video description.\\
    \hspace{6mm}- Completeness: Check if the option addresses all aspects of the question without omitting important details.\\
    \hspace{3mm}- 5. Make a Decision: Based on the analysis, choose the option that most accurately and completely answers the question in relation to the video information.\\
    \hspace{3mm}- 6. Justify Your Choice: Be prepared to explain why this option is the best choice, citing specific instances from the video information and how the option addresses the question comprehensively.\\
\#\# Input:\\
\hspace{3mm}\textcolor{DeepBlue}{- Video\_info:}\\
\hspace{6mm}\textcolor{DeepBlue}{- video theme: Make a Plaster Mask}\\
\hspace{6mm}\textcolor{DeepBlue}{- duration: 103.0s}\\
\hspace{6mm}\textcolor{DeepBlue}{- key action within specific time intervals:}\\
\hspace{9mm}\textcolor{DeepBlue}{- 35s $\sim$ 48s : clean out the face}\\
\hspace{9mm}\textcolor{DeepBlue}{- 48s $\sim$ 60s : apply tissue using water on face}\\
\hspace{9mm}\textcolor{DeepBlue}{- 60s $\sim$ 72s : apply it for full face}\\
\hspace{9mm}\textcolor{DeepBlue}{- 72s $\sim$ 77s : put it under the neck}\\
\hspace{9mm}\textcolor{DeepBlue}{- 77s $\sim$ 79s : dry it out}\\
\hspace{3mm}\textcolor{DeepBlue}{- Question: What is the process that takes place after cleaning out the face?}\\
\hspace{3mm}\textcolor{DeepBlue}{- Options:}\\
\hspace{6mm}\textcolor{DeepBlue}{A. After cleaning the face, the mask is dried out immediately.}\\
\hspace{6mm}\textcolor{DeepBlue}{B. After cleaning the face, tissue is applied on the face using water.}\\
\hspace{6mm}\textcolor{DeepBlue}{C. After cleaning the face, the plaster mask is removed and cleaned.}\\
\hspace{6mm}\textcolor{DeepBlue}{D. Once the face is clean, a special plaster is applied straight away.}\\
\hspace{6mm}\textcolor{DeepBlue}{E. Following the face cleaning, the plaster mask is put under the neck.}\\
\#\# Output:\\
    \hspace{3mm}- After completing the analysis, state your chosen option, including both the option letter and its content, and provide a detailed justification for your choice.\\
    \hspace{3mm}- Please structure all of your response in the following JSON format: \{\texttt{"Best\_option"}: xxx, \texttt{"Reason"}: xxx\}. Ensure that the response is structured correctly in JSON format, as it will be parsed using \texttt{json.loads()} in Python.\\
    \end{tabular}
\end{tcolorbox}
\caption{Prompt Template for Choosing The Best Option With Video and Enhancing Options' Distractors. The content highlighted in \textcolor{DeepBlue}{blue} changes according to the specific Q\&A.}
\label{tab:Prompt Template for Choosing The Best Option With Video and Enhancing Options' Distractors}
\end{minipage}
\end{table*}

\begin{table*}[t!]\centering
\begin{minipage}{0.95\textwidth}
\centering
\begin{tcolorbox} 
    \centering
      \small
    \begin{tabular}{p{0.95\textwidth}}
   \textcolor{blue}{\textbf{System message}} \\
You are tasked with evaluating whether a student has correctly answered a question based solely on the provided options.\\
    \midrule
   \textcolor{blue}{\textbf{Prompt}} \\
\#\# Instruction:\\
    \hspace{3mm}- 1. Understand the Question: Clearly grasp the question being asked, focusing on the specific information or decision that needs to be addressed.\\
    \hspace{3mm}- 2. Analyze User's Choice: Compare the user's selected option with the correct answer or the most appropriate response based on the information provided in the question.\\
    \hspace{3mm}- 3. Evaluate Correctness:\\
    \hspace{6mm}- Accuracy: Check if the user's choice accurately reflects the information provided in the question.\\
    \hspace{6mm}- Relevance: Assess whether the selected option directly answers the question, considering the details described.\\
    \hspace{3mm}- 4. Make a Judgement: Decide whether the user's choice is correct based on the alignment with the question's requirements.\\
    \hspace{3mm}- 5. Provide Feedback: If the choice is incorrect, offer constructive feedback on why it was not the best option and suggest how to improve the decision-making process for future questions.\\
\#\# Input:\\
\hspace{3mm}\textcolor{DeepBlue}{- Question: What is the process that takes place after cleaning out the face?}\\
\hspace{3mm}\textcolor{DeepBlue}{- Options:}\\
\hspace{6mm}\textcolor{DeepBlue}{A. After cleaning the face, one starts to make the mask shape.}\\
\hspace{6mm}\textcolor{DeepBlue}{B. After cleaning the face, tissue is applied on the face using water.}\\
\hspace{6mm}\textcolor{DeepBlue}{C. After cleaning out the face, the mask is dried out.}\\
\hspace{6mm}\textcolor{DeepBlue}{D. After cleaning the face, it is placed under the neck.}\\
\hspace{6mm}\textcolor{DeepBlue}{E. After cleaning the face, the mask needs to be left still for a few seconds.}\\
\hspace{3mm}\textcolor{DeepBlue}{- Correct Answer: B. After cleaning the face, tissue is applied on the face using water.}\\
\hspace{3mm}\textcolor{DeepBlue}{- Student's Answer: E. After cleaning the face, the mask needs to be left still for a few seconds.}\\
\#\# Output:\\
    \hspace{3mm}- If the student's answer is correct, please return the judgement with 'Yes', otherwise 'No'. And give the 'Reason' for the judgement.\\
    \hspace{3mm}- Please structure all of your response in the following JSON format: \{\texttt{"Judgement"}: xxx, \texttt{"Reason"}: xxx\}. Ensure that the response is structured correctly in JSON format, as it will be parsed using \texttt{json.loads()} in Python.\\
    \end{tabular}
\end{tcolorbox}
\caption{Prompt Template for Checking Consistency Between Predicted Option and Correct Option. The content highlighted in \textcolor{DeepBlue}{blue} changes according to the specific Q\&A.}
\label{tab:Prompt Template for Checking Consistency Between Predicted Option and Correct Option}
\end{minipage}
\end{table*}

\begin{table}[t!]\centering
\begin{minipage}{0.95\textwidth}
\centering
\begin{tcolorbox} 
    \centering
      \small
    \begin{tabular}{p{0.95\textwidth}}
   \textcolor{blue}{\textbf{System message}} \\
You are an advanced AI model specialized in analyzing video content. You will be given a series of frames extracted from a video. Your task is to carefully watch these frames, paying close attention to the sequence of events, object details. Based on your observations, answer the subsequent questions accurately.\\
    \midrule
   \textcolor{blue}{\textbf{Prompt}} \\
\#\# Instruction:\\
    \hspace{3mm}- 1. Observe Frames: Carefully analyze the sequence of frames provided, noting key events, object details, and any changes over time.\\
    \hspace{3mm}- 2. Understand the Question: Clearly grasp the question being asked, focusing on the specific information or decision that needs to be addressed based on the video content.\\
    \hspace{3mm}- 3. Analyze Options: Compare each option with the observations made from the frames, ensuring that the chosen option aligns with the details in the video.\\
    \hspace{3mm}- 4. Select the Correct Answer: Choose the option that best answers the question based on the analysis of the frames.\\
    \hspace{3mm}- 5. Provide the Answer: Answer with the option’s letter from the given choices directly.\\
\#\# Input:\\
\hspace{3mm}\textcolor{DeepBlue}{- Given Frames: video\_frame-0, ......, video\_frame-i}\\
\hspace{3mm}\textcolor{DeepBlue}{- Question: xxx}\\
\hspace{3mm}\textcolor{DeepBlue}{- Options:}\\
\hspace{6mm}\textcolor{DeepBlue}{A. Option A}\\
\hspace{6mm}\textcolor{DeepBlue}{B. Option B}\\
\hspace{6mm}\textcolor{DeepBlue}{C. Option C}\\
\hspace{6mm}\textcolor{DeepBlue}{D. Option D}\\
\hspace{6mm}\textcolor{DeepBlue}{E. Option E}\\
\#\# Output:\\
    \hspace{3mm}- Answer with the option’s letter from the given choices directly.\\
    \end{tabular}
\end{tcolorbox}
\caption{Prompt Template for Analyzing Video Content and Answering Questions. The content highlighted in \textcolor{DeepBlue}{blue} changes according to the specific Q\&A.}
\label{tab:Prompt Template for Analyzing Video Content and Answering Questions}
\end{minipage}
\end{table}

\clearpage
\bibliography{references}